# Efficient auto-labeling of large-scale poultry datasets (ALPD) using an ensemble model with self- and active-learning approaches

R.B. Bist[a], L. Chai[b], S. Weimer[c], H. Atungulu[d], C. Pennicott[c], X. Yang[b], C. Pallerla[e], Y. Tian[a], D. Wang[a,e,*]

[a] *Department of Biological and Agricultural Engineering, University of Arkansas, Fayetteville, AR, USA, 72701*

[b] *Department of Poultry Science, University of Georgia, Athens, GA, USA, 30602*

[c] *Department of Poultry Science, University of Arkansas, Fayetteville, AR, USA, 72701*

[d] *Haas Hall Academy, Springdale, AR, USA, 72764*

[e] *Department of Food Science, University of Arkansas, Fayetteville, AR, USA, 72704*

* *Corresponding: dongyiw@uark.edu*

**Abstract**

The rapid growth of artificial intelligence in poultry farming has highlighted the challenge of efficiently labeling large, diverse datasets. Manual annotation is time-consuming and costly, making it impractical for modern systems that continuously generate data. This study addresses this challenge by exploring semi-supervised auto-labeling methods, integrating self and active learning approaches to develop an efficient, label-scarce framework for auto-labeling large poultry datasets (ALPD). For this study, video data were collected from broilers and laying hens housed. Various machine learning models, including zero-shot models and supervised models were utilized for broilers and hens' detection. The results showed that YOLOv8s-World and YOLOv9s performed better when compared performance metrics for broiler and hen detection under supervised learning, while among the semi-supervised model, YOLOv8s-ALPD achieved the highest precision (96.1%) and recall (99%) with an RMSE of 1.87. The hybrid YOLO-World model, incorporating the optimal YOLOv8s backbone with zero-shot models, demonstrated the highest overall performance. It achieved a precision of 99.2%, recall of 99.4%, and an F1 score of 98.7% for detection. In addition, the semi-supervised models with minimal human intervention (active learning) reduced annotation time by over 80% compared to full manual labeling. Moreover, integrating zero-shot models with the best models enhanced broiler and hen detection, achieving comparable results to supervised models while significantly increasing speed. In conclusion, integrating semi-supervised auto-labeling and zero-shot models significantly improves detection accuracy. It reduces manual annotation efforts, offering a promising solution to optimize AI-driven systems in poultry farming, advancing precision livestock management, and promoting more sustainable practices.

## 1. Introduction

In recent years, the livestock and poultry sectors have faced significant challenges due to rising labor costs and persistent labor shortages, particularly during peak seasons. In 2024, labor expenses, including both cash and noncash employee compensation, are projected to increase by $3.0 billion (6.1 percent), reaching $52.0 billion compared to 2023 (USDA, 2024). Additionally, livestock and poultry purchases are expected to rise by $4.4 billion (10.2 percent), totaling $47.4 billion. These trends are expected to persist, emphasizing the urgent need for advanced technologies like artificial intelligence (AI) and machine learning (ML) to enhance automation, decision-making, and efficiency in agricultural systems (Shaikh et al., 2022). These AI-driven systems rely heavily on high-quality labeled datasets, which are the foundation for training robust models. Available models are often limited to basic detection, segmentation, classifications, and auto-labeling like "chicken" or “birds.” However, it cannot distinguish between breeds, their behaviors, or physical conditions. Poultry data varies significantly in breed, behavior, appearance, and rearing conditions. This diversity makes accurate and timely annotation crucial for quality control, behavior and welfare analysis, health monitoring, and process optimization.

Deep learning has significantly advanced automatic object detection, yet the need for large, high-quality, labeled datasets often constrains its practical application. Current approaches, particularly those based on supervised training, rely heavily on these datasets, posing a major challenge for widespread implementation. Supervised methods are inherently data-intensive and depend heavily on the availability of large, high-quality labeled datasets (Cao, 2024). Even with expert annotations, label noise or errors due to annotator variation are inevitable. Fully supervised labeling is effective but demands extensive manual annotation of large datasets, making it both time-consuming and expensive (Sapkota et al., 2024). This approach is often impractical in the poultry sector, where data is continuously generated in vast quantities, and tasks such as detecting subtle quality variations in poultry products require expert knowledge, adding to the complexity. Previous research on supervised learning models for object detection, which rely on paired images and labels, often faces limitations based on the volume and quality of annotated data (Wang et al., 2022). While large and diverse datasets enhance a model's pattern recognition and generalization to new images (Alzubaidi et al., 2023), creating such datasets is labor-intensive, potentially hindering the adoption of deep learning methods. To overcome this, semi-supervised learning, unsupervised domain adaptation, and noisy label learning have emerged as key methods to handle imperfect datasets, but each faces its own limitations, such as inefficiencies in filtering noisy labels and inadequate handling of limited data (Chen et al., 2022). Self-supervised deep learning and transformer architectures are to develop automatic labeling method that generates bounding box annotations for images with multiple objects leveraging both labeled and unlabeled data, thereby reducing the need for extensive manual labeling (Wang et al., 2022). These semi-supervised methods can guide the labeling process, improving accuracy while reducing the annotation burden (Adsule et al., 2024). Inductive semi-supervised methods, like self-training, start with a supervised model trained on labeled data and generate pseudo-labels for unlabeled data. These pseudo-labeled samples, along with the original labeled data, iteratively retrain the model to refine its performance. Active learning further complements this process by identifying the most informative samples from human intervention, ensuring that labeling efforts focus on data points that maximize the model's learning potential (Qazi et al., 2024). This hybrid approach of machine labeling augmented by human corrections addresses the limitations of manual annotation, saving time and resources while ensuring data quality and reliability. By improving the efficiency of data labeling, these techniques enhance the performance of AI-driven systems, particularly in agricultural applications like behavior monitoring and health assessment in poultry.

Unsupervised learning models are required to solve labeling problems, which operate without labeled data. These models can automatically label all the images as they are trained in a large dataset but lack training in the specific domain such as classifying all poultry species. In addition, it often struggles to achieve the accuracy required for critical applications due to the inherent noise and variability in agricultural data (Usama et al., 2019). Automatically generated labels are often considered "noisy" due to their lower annotation quality than manual labeling (Culman et al., 2023). To address this issue, a study pre-trained an object detection model using these noisy labels and fine-tuned the model with a limited amount of manually labeled data (Ahn et al., 2023; Culman et al., 2023). Recent advancements have demonstrated that combining noisy labels with a limited amount of manually labeled data can significantly enhance the detection precision, with gains of up to 37.0% compared to models trained solely on limited labeled data (Culman et al., 2023). Precision gains were also observed on unseen images, ranging from 17.6% to 38.5%, depending on the camera setup. These findings underscore the effectiveness of automatic labeling in scenarios with insufficient training data, where manual labeling is constrained by resources or time.

Current research in auto-labeling (**Fig. 1**) within agriculture has primarily explored methods like self-training and active learning, with promising results in domains such as image classification and segmentation (Li et al., 2023). Despite the progress, the application of these methods in the poultry industry remains underexplored, presenting a significant opportunity for innovation. Self-training, for instance, uses confidence thresholds to include only high-confidence pseudo-labeled samples in the retraining process, which minimizes errors and enhances model reliability (Wang et al., 2021). This research also adopts a self-training approach, augmented by human intervention to relabel incorrect or ambiguous images, thereby combining the strengths of automated and manual annotation. Moreover, this study integrates a zero-shot detection framework with the trained model to further improve accuracy and facilitate targeted image retrieval. This novel combination allows producers to locate specific birds or conditions through simple text inputs, providing an intuitive and efficient solution for monitoring and managing poultry datasets. These advancements highlight the potential of semi-supervised learning methods to address the challenges of data annotation in agriculture, paving the way for more scalable and precise AI applications.

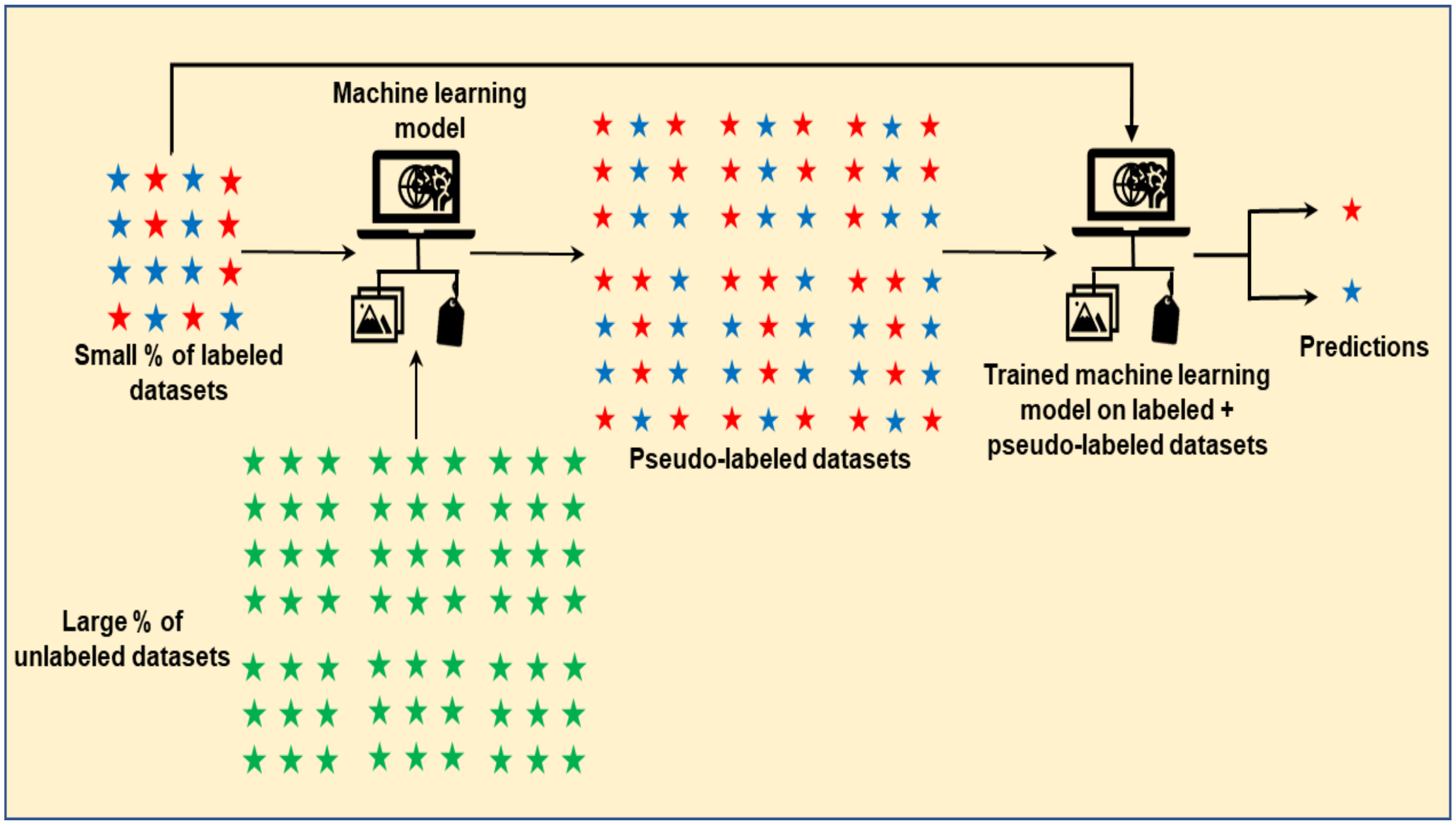

**Fig. 1.** Example of the semi-supervised learning model.

This study explored label-efficient learning methods within deep learning and their poultry applications, highlighting their significance in overcoming data annotation challenges in poultry farming. The objectives of this study were to (a) test and compare different zero-shot, supervised, and semi-supervised auto-labeling models, (b) develop and evaluate a semi-supervised auto-labeling method tailored to the unique challenges of the poultry industry, particularly in the context of large, diverse datasets, and (c) investigate the effectiveness of combining best model with zero-shot detector to improve detection and auto-labeling of broilers and layers. By achieving these objectives, this research aims to significantly advance the development of robust training datasets, enabling producers to effortlessly detect and monitor various individual birds in poultry through intuitive text-based inputs. This study not only addresses the pressing challenges of data annotation in agriculture but also contributes to advancing AI-driven solutions, fostering greater efficiency, scalability, and precision in the poultry industry.

## 2. Materials and Methods

The Institutional Animal Care and Use Committee (IACUC) reviewed and approved this study involving broilers under protocol #23009 at the University of Arkansas. Additionally, a related laying hen study was approved by the IACUC under protocol AUP# A2020 08-014-A2 at the University of Georgia.

### *2.1. Housing and management*

#### 2.1.1. Broilers

The study was conducted in broiler houses designed with specific configurations and management practices. Each house comprised 12 pens measuring 3.66m (12ft) x 2.44m (8ft), with individual programmable environmentally controlled research chambers utilized to maintain precise environmental conditions; average daily temperatures and relative humidity levels were monitored within these chambers and placed as per Cobb 500 broiler management guidelines. The experimental design included raising Cobb 500 male broilers, with a stocking density of $1.92ft^2$ per bird, and each pen housing 50 broilers aged from day 0 to 42. This study was conducted from February 1, 2023, to March 15, 2023. No treatments were applied during the study period, and the broilers received only commercial starter, grower, and finisher diets without additional enzymes or medications. For monitoring, cameras were strategically installed at a height of 2m (6.5ft), with angles perpendicular to the ground, ensuring comprehensive coverage of feeding stations and water sources within each pen. The camera setup utilized a LOREX 4K 32ch 8TB Wired NVR System (Model # N883A38B, Lorex Corporation, Markham, ON, Canada) with dual 8TB internal hard drives for recording, capturing footage at a frame rate of 30 Frame Per Second (FPS) and a resolution of 1920 x 1080.

2.1.2. Laying hens

The experiment was conducted in four identical research houses, each housing 200 Hy-Line W-36 hens. Birds were raised from day 1 to day 630 (90 weeks) in rooms measuring 7.3 m × 6.1 m × 3 m, equipped with perches, litter flooring, feeders, drinkers, lights, and nest boxes. Pine shavings were used as bedding, and a Chore-Tronics Model 8 controller (Chore-Time Equipment, Milford, IN) controlled environmental conditions such as temperature, humidity, light, and ventilation. Behavior was recorded using six night-vision network cameras (PRO-1080MSB, Swann Communications USA Inc., Santa Fe Springs, LA) mounted 3 meters above the litter floor and two cameras 0.5 meters above the ground, capturing footage 24 hours videos a day. Videos were stored in .avi format at a resolution of 1920 × 1080 pixels and 15 frames per second, using a digital video recorder (DVR-4580, Swann Communications USA Inc., Santa Fe Springs, LA) for data storage, covering the period from 25 to 50 weeks of age. The previously published paper explained the details of housing and management (Bist et al., 2023; X. Yang et al., 2023).

*2.2. Image labeling and data pre-processing*

The images collected from broiler and laying hen experimental rooms were first preprocessed and filtered to ensure they met the requirements for training a machine learning model. The dataset was then divided into training (60%), validation (20%), and testing (20%) subsets. To enhance model performance and generalization, various augmentation techniques, including geometric transformations (such as rotations, flipping, and scaling), were applied following the previous paper (Bist et al., 2024). These augmentations simulate real-world variations, improve model robustness, prevent overfitting, and enhance performance on unseen data. The final dataset distributions after augmentation are summarized in **Tables 1** and **Table 2**. These tables present the labeled training and validation datasets along with the unlabeled test datasets. Besides unlabeled test datasets, this study also has other unlabeled image categories, often referred to as pseudo-label images, which were later assigned pseudo-labels using the best-performing model from training. **Table 1** was used to evaluate model performance and compare confidence scores. Additionally, **Table 2** highlights the varying quantities of images used to determine the minimum required for optimal results. Similarly, the Supervised I-640 dataset was used for model comparisons on behavioral data.

**Table 1.** Datasets distribution for broilers and laying hens used in this experiment to compare different models' performance.

| Subsets | Broilers | | | | | Laying hens | | | | |
|---|---|---|---|---|---|---|---|---|---|---|
| | **Train** | **Val** | **Test** | **Pseudo-label Images** | **Total** | **Train** | **Val** | **Test** | **Pseudo-labels Images** | **Total** |
| Images | 120 | 40 | 40 | 1500 | 1700 | 120 | 40 | 40 | 1500 | 1700 |
| Instances | 5437 | 1830 | 1853 | 70,251 | 79,371 | 3981 | 1356 | 1634 | 48,428 | 55,399 |

**Table 2.** Different number of datasets were used to compare model auto-labeling performance.

| Images | Subsets | Broilers | | | | Laying hens | | | |
|---|---|---|---|---|---|---|---|---|---|
| | | **Train** | **Val** | **Test** | **Total** | **Train** | **Val** | **Test** | **Total** |
| **I50** | Images | 15 | 5 | 5 | 25 | 15 | 5 | 5 | 25 |
| | Instances | 701 | 232 | 231 | 1,164 | 643 | 342 | 265 | 1,250 |
| **I100** | Images | 30 | 10 | 10 | 50 | 30 | 10 | 10 | 50 |
| | Instances | 1422 | 399 | 461 | 2,282 | 1193 | 605 | 439 | 2,237 |
| **I200** | Images | 60 | 20 | 20 | 100 | 60 | 20 | 20 | 100 |
| | Instances | 2824 | 864 | 927 | 4,615 | 2293 | 1022 | 813 | 4,128 |
| **I400** | Images | 120 | 40 | 40 | 200 | 120 | 40 | 40 | 200 |
| | Instances | 5437 | 1803 | 1853 | 9,120 | 3981 | 1356 | 1634 | 6,971 |

Where I50, I100, I200, and I400 are the image quantity 50, 100, 200, and 400, respectively.

For labeling, we utilized tools including MakeSense.ai and CVAT to annotate the training, validation, and initially unlabeled datasets. Trained specialists in manual labeling were hired and cross-checked their labeled datasets by the author before training the models with the help of given tools. The labeled datasets were then exported in YOLO text and Pascal VOC formats to ensure compatibility with different machine learning frameworks. These labeled datasets were subsequently used to train various machine-learning models. The performance of the trained models was evaluated using validation and test datasets to determine their effectiveness. The best-performing model was then employed to automatically label the remaining unlabeled images, generating pseudo-labeled data.

The following experimental setup in **Table 3** was used for model training and evaluation, designed to ensure optimal performance and efficient training. This setup included high-performance hardware, software, and training configuration. The system utilized an NVIDIA A100 GPU with CUDA acceleration, Python-based tools, and key machine-learning libraries to facilitate the training and evaluation of the models provided by the Arkansas High-Performance Computing Center located at the University of Arkansas.

**Table 3.** Experimental setup utilized for model evaluation.

| **Configuration** | **Parameters** |
|---|---|
| Operating system | Linux-Based System |
| Accelerated environment | NVIDIA CUDA |
| GPU | A100, 64 Cores, 40 GB VRAM |
| VRAM | 40 GB |
| Computer CPU | Intel® Core™ Ultra 7 155H 3.80GHz |
| Computer RAM | 32 GB |
| Software tool | Jupyter Notebook |
| Python | Python 3.11.5 |
| Libraries | Tensorflow-2.18.0, Torch 2.5.1+cu124, Torch-vision 0.20.1+cu124, CUDA version 12.4, NumPy 2.1.3 |
| Training setup | Batch size of 16, learning rate of 0.001, 300 epochs |

### *2.3. Model description*

In this study, we utilized various machine learning models for auto-labeling to efficiently generate labeled data for model training. The models employed for this purpose span different categories, including zero-shot, supervised, and semi-supervised models, each with distinct methodologies and capabilities.

### 2.3.1. Zero-shot models for auto-labeling

#### 2.3.1.1. Grounding DINO (GDINO)

Grounding DINO is a state-of-the-art zero-shot object detection model developed by IDEA-Research (S. Liu et al., 2025), designed to detect arbitrary objects or behaviors based on human input, such as category names or referring expressions. Built on the DINO model, a transformer-based architecture known for its success in image classification and object detection, Grounding DINO incorporates a grounding module. This addition enables the model to establish a relationship between language and visual content by learning to associate words in text descriptions with specific regions in images. As a result, Grounding DINO can detect objects in unseen images without prior knowledge of those objects, offering high versatility for a range of real-world applications, including agriculture.

Key components of Grounding DINO include the image and text backbones, which extract features from input images and corresponding text descriptions, respectively (Liu et al., 2025). The feature enhancer fuses these modalities, allowing for cross-modality information exchange, while the language-guided query selection initializes queries using language inputs. Finally, the cross-modality decoder predicts bounding boxes based on the fused features and queries, enabling the model to localize and identify objects accurately. The model's ability to generalize to new objects without requiring additional training data makes it an efficient and adaptive solution for zero-shot detection. In this configuration, we fine-tune the initial GDINO model on a small, labeled dataset to improve its precision and recall for poultry behavior detection. Fine-tuning enhances the model's ability to predict more accurate labels when applied to unlabeled images. Non-maximum suppression is applied to remove duplicate detections. This technique calculates the overlap between detected objects using the Intersection over Union (IoU) metric. If the overlap exceeds a specified threshold, the detections are considered redundant. These duplicates are eliminated starting with the detection with the lowest confidence score. The threshold value ranges from 0 to 1, with smaller values making non-maximum suppression more selective in removing overlapping detections.

2.3.1.2. YOLO-World

YOLO-World is an advanced extension of the YOLO framework, designed by AI Lab to address the limitations of traditional object detection models that rely on predefined categories (Cheng et al., 2024). By incorporating vision-language modeling and pre-training on large-scale datasets, YOLO-World enables open-vocabulary detection, allowing it to identify a diverse range of objects in a zero-shot manner without requiring task-specific annotations. The model features a Re-parameterizable Vision-Language Path Aggregation Network (RepVL-PAN), which facilitates effective interaction between visual and linguistic information and utilizes a region-text contrastive loss to enhance alignment between image and text data. YOLO-World has demonstrated superior performance on challenging benchmarks, such as the LVIS dataset, achieving 35.4 Average Precision (AP) with 52.0 Frames Per Second (FPS) on a V100 GPU, surpassing several state-of-the-art methods in both accuracy and speed. Its ability to generalize to unseen categories is further proven through its fine-tuned performance on downstream tasks like object detection and open-vocabulary instance segmentation. The integration of transfer learning and pre-trained weights allows YOLO-World to analyze visual data for tasks without relying on annotated data, and its architecture, which supports user-defined vocabulary prompts, makes it adaptable to various detection tasks. This versatility ensures rapid and accurate results across diverse applications, including poultry behavior analysis in agriculture.

#### 2.3.1.3. CLIP

CLIP (Contrastive Language-Image Pre-Training) is an advanced neural network developed by OpenAI, designed to bridge the gap between visual and textual information (OpenAI 2024). Trained on a diverse dataset of image-text pairs, CLIP predicts a given image's most relevant textual description through natural language instructions, demonstrating zero-shot learning capabilities similar to models like GPT-2 and GPT-3. Remarkably, CLIP achieves comparable performance to ResNet50 on the ImageNet "zero-shot" benchmark without relying on the 1.28 million labeled examples from the original dataset (OpenAI 2024). This achievement highlights its ability to address critical challenges in computer vision, such as dependency on large, labeled datasets and generalization to novel tasks. By aligning visual and textual modalities, CLIP has proven highly versatile for applications requiring a deep understanding of image-text relationships, including image captioning, visual question answering, and object detection.

This study employed CLIP to align image regions with textual descriptions using cosine similarity between image and text embeddings (Wolff et al., 2023). For a batch of N image-text pairs, the model calculates similarity scores for all possible combinations, identifying correct pairings by maximizing the similarity between matched embeddings and minimizing it for mismatched pairs. Through joint training of image and text encoders, CLIP creates a shared embedding space that facilitates accurate mapping of textual labels to image regions. This capability is particularly valuable in dynamic and diverse agricultural environments, where it enhances tasks, such as behavior detection, annotation, and interpretation. By leveraging CLIP's matrix-based representation, the study effectively improved the understanding and integration of visual and textual data, advancing applications in precision agriculture and beyond (Wolff et al., 2023).

### 2.3.2. Supervised models for auto-labeling

#### 2.3.2.1. YOLO models

For this research, we utilized multiple versions of the YOLO (You Only Look Once) detector series by Ultralytics, including YOLOv8s, YOLOv9s, YOLOv10s, YOLO11s, YOLOv8s-Worldv2, and RT-DETR (Ultralytics, 2024). These models were selected for their efficiency and performance in object detection tasks focusing on their suitability for auto-labeling and detection in poultry-related applications. YOLOv8s and other small variants were chosen for their speed and lower resource consumption than larger models, making them ideal for real-time processing and handling large datasets with reduced computational overhead. While larger models provide higher accuracy, the smaller YOLO models balance speed and performance, enabling rapid detection and labeling without requiring extensive hardware resources. YOLOv8s-Worldv2, a more advanced version, was specifically leveraged for its open-vocabulary detection capabilities, allowing greater flexibility in identifying a wide range of objects without predefined labels. This combination of models provides a versatile, scalable solution for efficient and accurate auto-labeling in the context of poultry behavior detection. Once trained, they are deployed to auto-label additional unlabeled images, creating pseudo-labeled data that can be used for further training iterations.

#### 2.3.2.2. Faster-RCNN

Faster-RCNN, a region-based convolutional neural network (CNN), was utilized for object detection in this study and analyzed using open-source GitHub repository (Ren et al., 2017; Yang et al., 2017). The model generates candidate regions within an image and subsequently classifies them into predefined categories. Initially, Faster-RCNN undergoes supervised training on a labeled dataset, allowing it to learn to detect various objects or behaviors. Once trained, the model can be applied to unlabeled images for auto-labeling tasks, particularly for identifying and classifying poultry behaviors. The ResNet backbone was employed for the feature extraction and image classification tasks due to its deep residual network architecture. ResNet's use of skip connections aids in effectively training deeper models, which is especially beneficial for accurately recognizing complex behaviors in poultry. After the model is trained on labeled datasets, it can generate pseudo-labels for various poultry behaviors, enabling efficient and scalable auto-labeling in real-world applications. In this research, Faster-RCNN with a ResNet backbone was selected for its robustness in object detection tasks and its ability to generate high-quality pseudo-labels for behavior recognition in poultry. This approach detects various behaviors without requiring extensive manual annotation, making it an effective tool for large-scale data labeling tasks.

### 2.3.3. Semi-supervised models for auto-labeling

#### 2.3.3.1. YOLO-ALPD

The YOLO models, including YOLOv8s, YOLOv9s, YOLOv10s, YOLOv11s, YOLOv8s-Worldv2 and RT-DETR, were initially trained on small, labeled datasets to develop highly accurate baseline models. These trained models were then employed to ALPD that were previously unlabeled. The ALPD approach integrates supervised learning with semi-supervised techniques to enhance model performance in data-scarce scenarios. Specifically, the models were first fine-tuned on the small labeled datasets to optimize detection accuracy. Subsequently, these models were applied to unlabeled datasets to generate pseudo-labels, which were iteratively incorporated back into the training process. During each iteration, the model's predictions on the unlabeled data were refined, creating a feedback loop that progressively improved its detection and labeling capabilities. This iterative auto-labeling strategy enabled the semi-supervised models to efficiently label large datasets while enhancing their performance, making them suitable for scalable and accurate detection tasks in poultry datasets.

#### 2.3.3.2. Faster-RCNN-ALPD

The Faster-RCNN with ResNet backbone models were adapted to employ the ALPD technique within a semi-supervised learning framework. Initially, these models underwent supervised training on small labeled datasets to establish a reliable baseline for detection tasks. Following this, the models were utilized to generate pseudo-labels for unlabeled poultry behavior datasets. These pseudo-labels were iteratively refined and incorporated back into the training process, creating a feedback loop that progressively enhanced model accuracy and generalization. In the case of Faster-RCNN, the model leveraged its region-based detection framework, while ResNet utilized its robust feature extraction capabilities to identify and label poultry datasets effectively. The ALPD approach facilitated large-scale data augmentation and reduced the dependency on extensive manual annotation, allowing the models to adapt to diverse datasets and improve their performance across successive iterations. This semi-supervised learning paradigm ensured continuous improvement, making these models highly effective for large-scale poultry dataset analysis.

#### 2.3.4. Final model architecture

The models employed for auto-labeling played a pivotal role in generating large-scale labeled datasets essential for training and enhancing machine learning models designed for poultry breed and behavior detection. This study utilized a combination of zero-shot learning, supervised learning, and semi-supervised learning with active learning techniques to develop a robust detection system with improved accuracy, while significantly reducing the reliance on manual labeling. Active learning helps the model selectively query the most informative data points for labeling to improve performance efficiently (Adsule et al., 2024). The approach facilitated scalable solutions for monitoring poultry breeds and behaviors, addressing key challenges in poultry management. To further enhance prediction capabilities, the best-performing model was integrated with zero-shot learning frameworks, incorporating text-based prompts to improve prediction accuracy and adaptability. This integration allowed the system to manage novel scenarios effectively and produce more detailed and comprehensive outputs, as illustrated in **Fig. 2.**

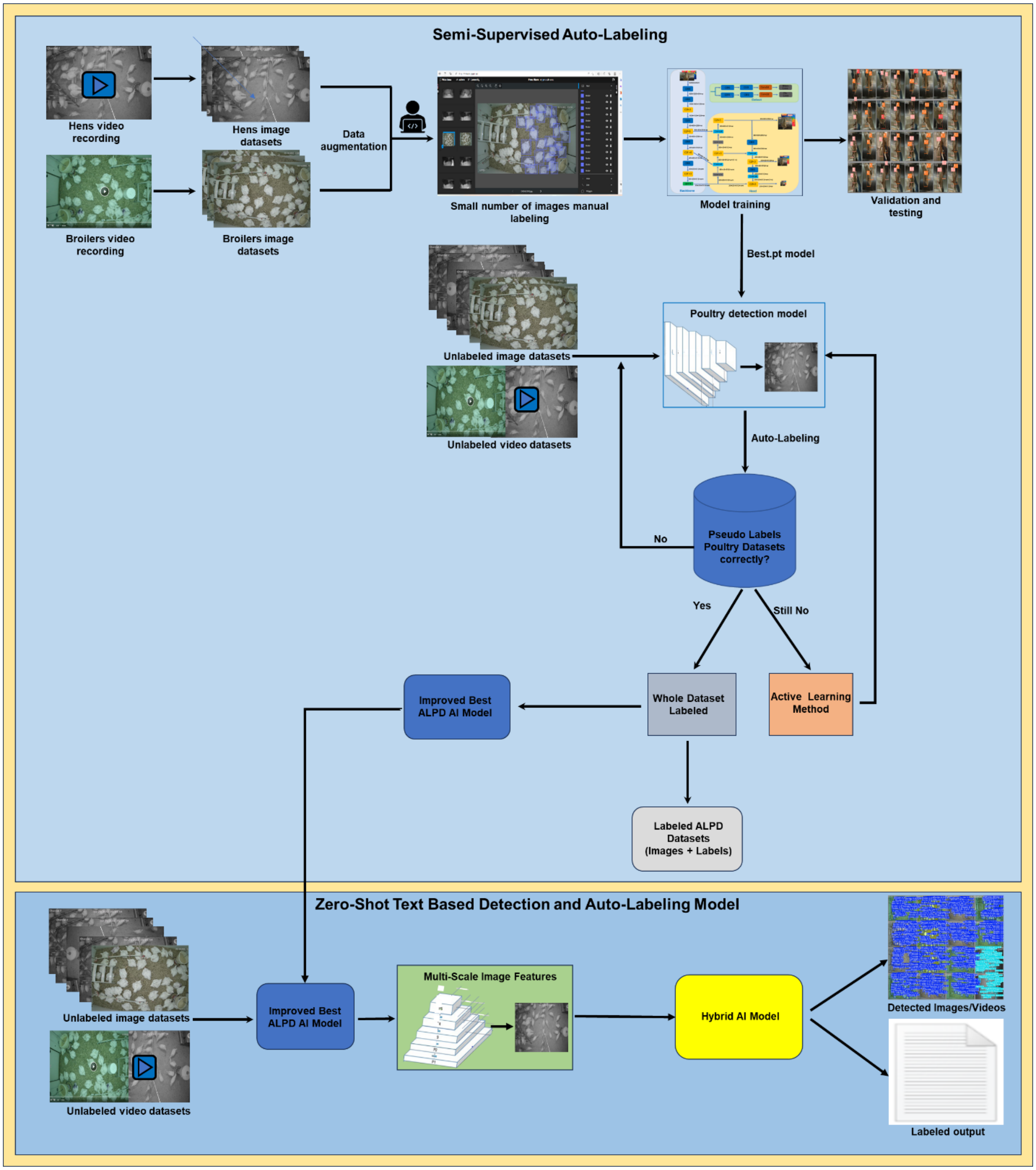

**Fig. 2.** Flowchart of the ALPD semi-supervised model integrated with a text-prompt auto-labeling and detection framework. Note: The initial phase employs a semi-supervised learning structure, which is subsequently combined with a fusion of semi-supervised learning and a zero-shot text-

based detection and auto-labeling approach. ALPD: Auto-Labeling Large Poultry Datasets; AI: Artificial Intelligence.

This study introduces a deep-learning framework called ALPD model, designed to address the challenges of semi-supervised learning, unsupervised domain adaptation, and noisy label learning in agricultural image detection. ALPD leverages models trained on limited annotated data (50 labeled images for each hen and broiler) or annotated source domain data to generate noisy labels for unlabeled datasets. A cross-model co-optimization approach is employed, where two networks are trained in parallel while exchanging information to filter and correct noisy labels through a cascaded process. This iterative refinement progressively improves label quality, offering a flexible and efficient solution for deep neural network models. ALPD model is particularly suited for large-scale, annotation-efficient tasks like automatic labeling in poultry datasets, addressing critical bottlenecks in data preparation for agricultural applications. To enhance annotation efficiency, the framework incorporates active learning, empowering labelers to focus on uncertain or critical data points that maximize model improvement.

### *2.3.5. Active learning framework*

This study employs four active learning frameworks to enhance model performance while minimizing manual labeling efforts: confidence thresholding, uncertain sample selection and labeling, query by committee, and model retraining with performance evaluation. These frameworks work together to iteratively improve the detection model by prioritizing uncertain and high-impact samples for labeling and retraining.

#### 2.3.5.1. Confidence thresholding

The Active Learning with Confidence Thresholding (ALCT) methodology aims to iteratively refine object detection performance. Initially, the model is trained on a small, labeled dataset to learn basic object representations. Once trained, the model makes predictions on the unlabeled data and assigns confidence scores to each detected object, reflecting the certainty of the predictions. A predefined confidence threshold is applied to these predictions, with any instances falling below the threshold flagged as uncertain (Wang et al., 2021). These uncertain samples are prioritized for relabeling again and returned to unlabeled images, ensuring the model focuses on the data points most likely to improve its performance. The new image samples are integrated into the training dataset, and the model is retrained. Over successive iterations, this process continually enhances the model's ability to make accurate predictions, improving its robustness and scalability for agricultural monitoring tasks.

2.3.5.2. Selection of uncertain samples and labeling

In this framework, the model ranks the unlabeled samples based on their confidence scores. Lower confidence scores indicate higher uncertainty, and these samples are extracted for further labeling (Wang et al., 2021). A user-defined confidence threshold, typically set below 50%, determines which samples are considered uncertain. These uncertain samples are then manually or semi-automatically labeled. In the semi-automated approach, the model generates initial labels, which are later verified and corrected by human annotators. This reduces the time required for manual annotation while ensuring high-quality labeling. Once labeled, the samples are added to the training dataset, enhancing its diversity and quality and improving the model's performance in subsequent iterations.

2.3.5.3. Query by committee

The query-by-committee approach identifies instances where the model's predictions conflict or exhibit significant uncertainty. In these cases, multiple models are used to generate different predictions, and the discrepancies between them guide human annotators to focus on clarifying these uncertainties. Resolving these conflicts enriches the dataset with critical, high-impact information, allowing the model to learn more effectively from ambiguous or complex examples.

2.3.5.4. Model retraining and performance evaluation

Following the incorporation of newly labeled samples, the model undergoes retraining to integrate the expanded dataset. This phase is essential for allowing the model to learn from the newly labeled data, enhancing its detection capabilities. During this process, the model adapts to the additional information, which helps refine its performance on previously unseen examples. Retraining involves fine-tuning the model on the updated dataset, optimizing the loss function, adjusting learning rates, and applying data augmentation techniques to improve generalization. After retraining, a separate validation dataset evaluates the model's performance. Key metrics, such as mean Average Precision (mAP), precision, recall, and F1-score, are used to assess the model's accuracy and its ability to generalize to unseen data. These evaluation results guide the iterative learning process, ensuring that each cycle leads to measurable improvements in model performance and efficiency.

## *2.4. Model evaluation metrics*

A range of commonly used evaluation metrics were employed to evaluate the models' performance for auto-labeling and pseudo-labeled data output (Cheng et al., 2024; Ultralytics, 2024). These metrics help assess the quality of the model's predictions, including its ability to detect behaviors and the accuracy of its auto-labeled outputs. The following metrics were applied:

### 2.4.1. Precision

Precision measures the proportion of correctly predicted positive instances out of all instances predicted as positive. It is crucial when false positives have a high cost.

$$Precision = \frac{TP}{TP + FP} \qquad (i)$$

Where, TP = True Positives and FP = False Positives

High precision ensures that the model accurately detects target behaviors, such as breeds or behaviors, without many false detections.

### 2.4.2. Recall

Recall, also known as sensitivity, measures the proportion of actual positive instances that were correctly identified by the model. It is especially useful when false negatives are costly.

$$Recall = \frac{TP}{TP + FN} \qquad (ii)$$

Where, FN = False Negatives

A high recall ensures that most of the true behaviors, such as breeds or behaviors, are identified, even if it results in some false positives.

### 2.4.3. Mean Average Precision

The mean average precision is a commonly used metric in object detection, evaluating the model's precision at a standard overlap threshold.

$$mAP = \frac{1}{N}\sum_{i=1}^{N} APi \qquad (iii)$$

Where, AP$i$ = Average Precision for class $i$ and N = Total number of classes

A higher mAP indicates better overall performance in detecting breeds or behaviors with an acceptable level of localization.

#### 2.4.4. F1-Score

The F1-score is the harmonic means of precision and recall, providing a single metric that balances both the ability to correctly identify positive instances and the ability to detect as many positives as possible.

$$F1 = 2 \times \frac{Precision \times Recall}{Precision + Recall} \qquad (iv)$$

The F1-score is particularly useful when there is an imbalance between classes (e.g., rare behaviors like mislaying). It provides a balanced measure of a model's performance, ensuring that both false positives and false negatives are accounted for in the evaluation.

#### 2.4.5. Intersection over Union (IoU)

IoU measures the overlap between the predicted bounding box and the ground truth bounding box. It is widely used to assess the localization accuracy in object detection tasks.

$$IoU = \frac{Area\ of\ overlap}{Area\ of\ union} \qquad (v)$$

A higher IoU indicates that the predicted bounding box closely matches the true position of the object (e.g., detected behavior). IoU is essential for understanding how well the model is localizing specific behaviors, such as mislaying or pecking.

#### 2.4.6. Mean Absolute Error (MAE)

MAE measures the average absolute difference between the true and predicted values, providing a straightforward indication of prediction accuracy.

$$MAE = \frac{1}{N}\sum_{i=1}^{N} |y_i - \hat{y}_i| \qquad (vi)$$

where $y_i$ is the true value and $\hat{y}_i$ is the predicted value. N refers to the total number of samples or data points in the evaluated dataset.

MAE measures the average magnitude of errors without considering their direction. Lower MAE values indicating better model performance.

2.4.7. Mean Squared Error (MSE)

MSE calculates the average squared difference between the true and predicted values, heavily penalizing larger errors.

$$MSE = \frac{1}{N}\sum_{i=1}^{N}(y_i - \hat{y}_i)^2 \qquad (vii)$$

MSE penalizes larger errors more heavily, making it sensitive to outliers. A lower MSE value signifies a model that makes fewer and smaller errors in its predictions.

2.4.8. Root Mean Square Error (RMSE)

RMSE is the square root of the MSE, offering a measure of prediction error in the same units as the data. Smaller RMSE values indicate better predictive accuracy.

$$RMSE = \sqrt{MSE} \qquad (viii)$$

2.4.9. False Negative Rate (FNR)

FNR represents the proportion of actual positive instances that were incorrectly classified as negative, indicating the model's failure to detect positives.

$$FNR = \frac{FN}{FN+TP} \qquad (ix)$$

where FN is the number of false negatives.

FNR quantifies the proportion of actual positives incorrectly identified as negatives.

## 3. Results and Discussions

### *3.1. Data distribution*

**Fig. 3** presents the distribution of instance labels across different datasets used for model training. However, while the images allocated for broilers and hens maintain the same quantity, the instances within each image vary. **Fig. 3a** shows the distribution of labels across the training datasets for broilers and hens, demonstrating a balanced representation of the various categories, ensuring the model encounters diverse instances during training. **Fig. 3b** displays the distribution of labels in the pseudo-label training datasets, where the model generates additional labels based on its predictions. While this expands the dataset, slight imbalances in certain categories can be observed due to the challenges in automatically labeling some instances. Overall, these distributions highlight the strengths of our approach in combining real and pseudo-labeled instances to create a robust training dataset.

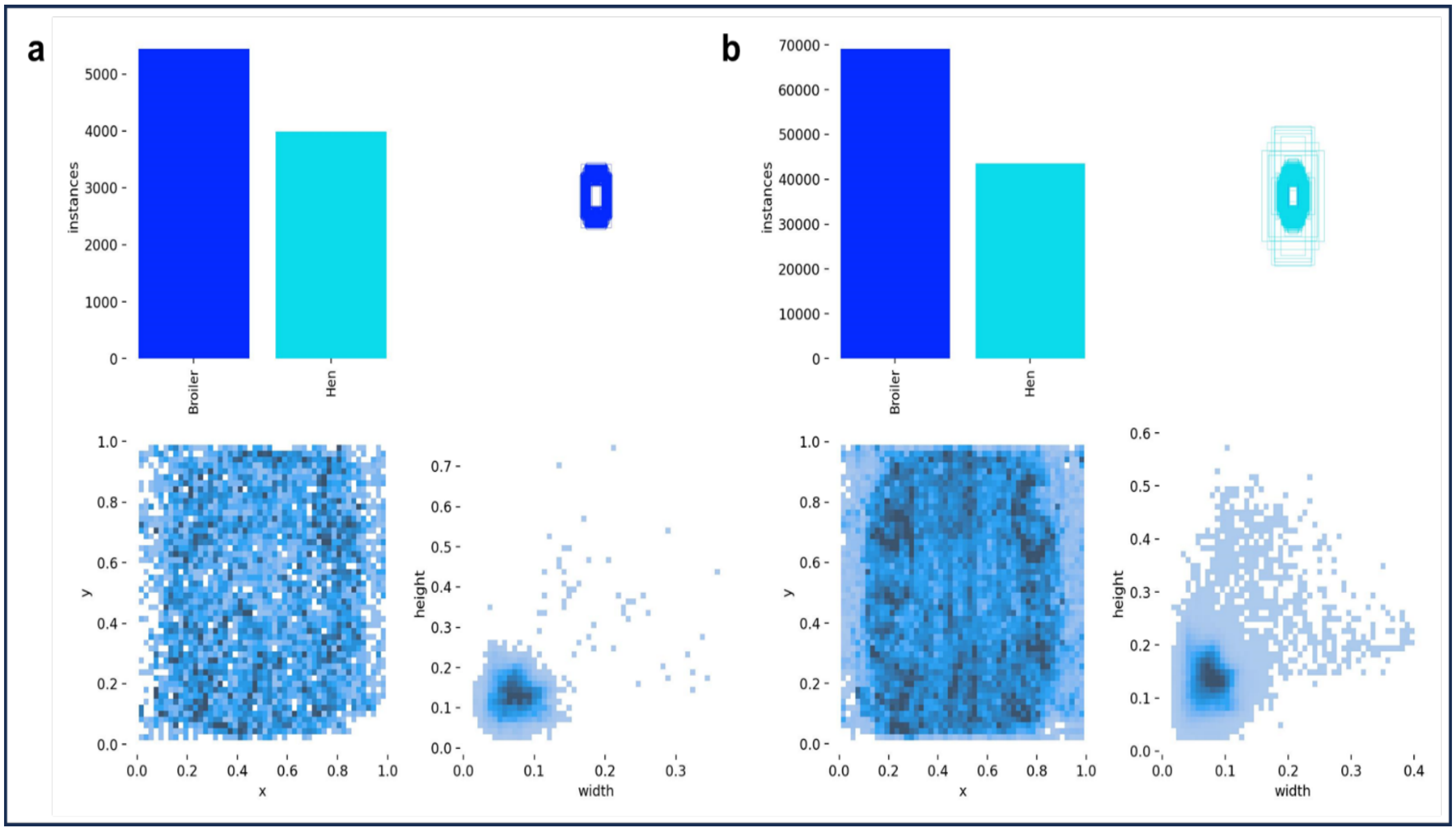

**Fig. 3.** Distribution of instances labels across a) Training datasets for broilers and hens, and b) Pseudo-label Datasets.

*3.2. Supervised and semi-supervised model comparison*

The performance metrics of supervised and semi-supervised models for broiler and hen detection reveal notable distinctions in their accuracy, precision, recall, and mAP scores across different models (**Table 4** and **Fig. 4**). Among the supervised models, YOLOv8s demonstrated higher performance for broiler detection with a precision of 97.3%, recall of 98.7%, and mAP_0.50 of 99.3%. This was closely followed by YOLOv9s, which achieved similarly high mAP_0.50 of 99.3% for broilers but with slightly better precision and recall values of 98.1% and 98.6%, respectively. Similarly, YOLOv9s for hen detection achieved the best performance with a recall of 91% and mAP_0.50 of 96.3%, outperforming YOLOv8s with a recall of 88.1% and mAP_0.50 of 96.1%. The YOLOv8s-World model, with an overall mAP_0.50 of 97.5%, also performed well across both breeds, demonstrating the model's adaptability to various conditions. However, it was outperformed by YOLOv9s in terms of precision and recall for broilers. The RT-DETR-l model, which represents a different architecture, displayed impressive performance for broiler detection (precision of 98.7%, recall of 98.2%) and hens' detection with recall at 93.4% and mAP_0.50 of 97.1%. In comparison, Faster-RCNN Restnet50 showed significantly lower performance across all categories, with a recall of just 62.8% and an overall mAP_0.50 of 94%. These results highlight the superior capabilities of the YOLO family and RT-DETR-based models in real-time object detection tasks for poultry, outperforming Faster-RCNN in terms of both precision and recall.

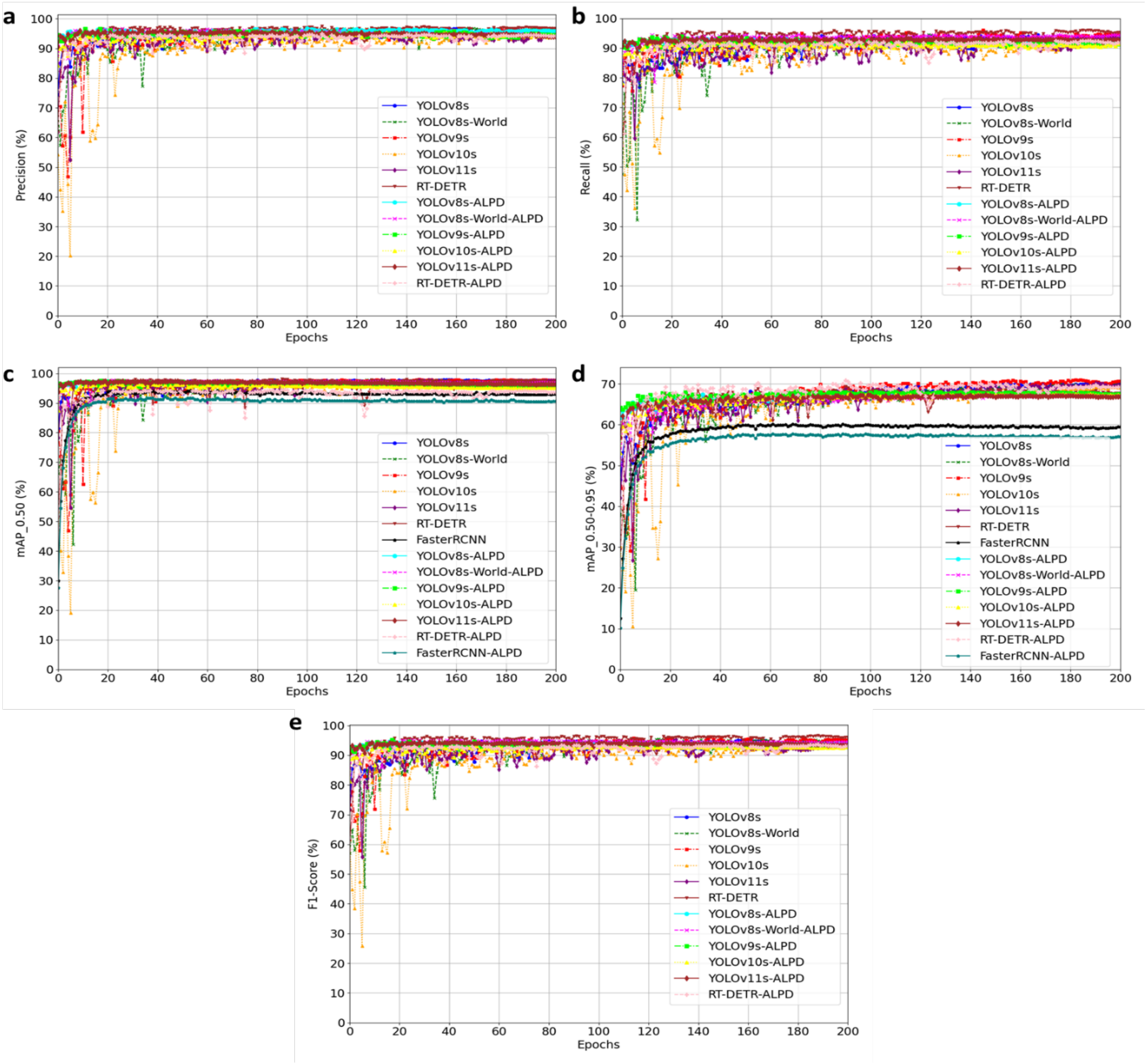

**Fig. 4.** Performance comparison of various models for broiler and hen detection with a) precision, b) recall, c) mAP_0.50, d) mAP_0.50-0.95, and e) F1-score. mAP is mean average precision; YOLO stands for You Only Look Once; ALPD stands for Auto-labeling of large poultry datasets; RT-DETR stands for Real-Time DEtector Transformer.

**Table 4.** Validation performance metrics of supervised and semi-supervised models for broiler and hen detection across various models.

| Categories | Models | Class | Precision | Recall | mAP_0.50 | mAP_0.50-0.95 | F1-score |
|---|---|---|---|---|---|---|---|
| Supervised | YOLOv8s | Broiler | 97.3 | 98.7 | 99.3 | 72.2 | 98.0 |
| | | Hen | 94.3 | 88.1 | 96.1 | 68.5 | 91.1 |
| | | Overall | 95.8 | 93.4 | 97.7 | 70.4 | 94.6 |
| | YOLOv8s-World | Broiler | 96.6 | 98.3 | 99.2 | 72.5 | 97.4 |
| | | Hen | 95.4 | 89.4 | 95.9 | 67.9 | 92.3 |
| | | Overall | 96.0 | 93.9 | 97.5 | 70.2 | 94.9 |
| | YOLOv9s | Broiler | 98.1 | 98.6 | 99.3 | 72.9 | 98.3 |
| | | Hen | 94.9 | 91.0 | 96.3 | 69.0 | 92.9 |
| | | Overall | 96.5 | 94.8 | 97.8 | 71.0 | 95.6 |
| | YOLOv10s | Broiler | 94.3 | 97.6 | 98.8 | 71.3 | 95.9 |
| | | Hen | 92.8 | 88.8 | 95.2 | 67.1 | 90.8 |
| | | Overall | 93.6 | 93.2 | 97.0 | 69.2 | 93.4 |
| | YOLOv11s | Broiler | 94.2 | 98.6 | 99.0 | 72.5 | 96.3 |
| | | Hen | 94.1 | 88.4 | 95.9 | 67.9 | 91.2 |
| | | Overall | 94.1 | 93.5 | 97.5 | 70.2 | 93.8 |
| | RT-DETR-l | Broiler | 98.7 | 98.2 | 99.1 | 72.1 | 98.4 |
| | | Hen | 95.0 | 93.4 | 97.1 | 66.7 | 94.2 |
| | | Overall | 96.7 | 95.8 | 98.1 | 69.4 | 96.2 |
| | Faster-RCNN Restnet50 | Broiler | 63.9 | 70.7 | 98.2 | 63.9 | 67.1 |
| | | Hen | 55.5 | 62.8 | 89.8 | 55.5 | 58.9 |
| | | Overall | 59.7 | 66.8 | 94.0 | 59.7 | 63.1 |
| Semi-supervised | YOLOv8s-ALPD | Broiler | 97.2 | 96.9 | 98.8 | 70.2 | 97.0 |
| | | Hen | 95.5 | 86.3 | 93.7 | 64.9 | 90.7 |

| | | | | | | |
|---|---|---|---|---|---|---|
| | **Overall** | 96.4 | 91.6 | 96.3 | 67.5 | 93.9 |
| **YOLOv8s-World-ALPD** | **Broiler** | 97.8 | 97.1 | 98.5 | 71.8 | 97.4 |
| | **Hen** | 93.5 | 91.2 | 95.2 | 67.5 | 92.3 |
| | **Overall** | 95.7 | 94.2 | 96.8 | 69.6 | 94.9 |
| **YOLOv9s-ALPD** | **Broiler** | 96.7 | 97.5 | 98.5 | 73.1 | 97.1 |
| | **Hen** | 95.1 | 87.3 | 92.9 | 68.0 | 91.0 |
| | **Overall** | 95.9 | 92.4 | 95.7 | 70.6 | 94.1 |
| **YOLOv10s-ALPD** | **Broiler** | 95.6 | 97.2 | 98.7 | 70.9 | 96.4 |
| | **Hen** | 93.0 | 85.5 | 93.4 | 64.1 | 89.1 |
| | **Overall** | 94.3 | 91.3 | 96.0 | 67.5 | 92.8 |
| **YOLOv11s-ALPD** | **Broiler** | 97.2 | 97.3 | 98.4 | 72.4 | 97.2 |
| | **Hen** | 92.4 | 89.7 | 93.6 | 66.5 | 91.0 |
| | **Overall** | 94.8 | 93.5 | 96.0 | 69.5 | 94.1 |
| **RT-DETR-l** | **Broiler** | 95.3 | 96.6 | 94.8 | 72.3 | 95.9 |
| | **Hen** | 93.8 | 87.8 | 92.2 | 67.2 | 90.7 |
| | **Overall** | 94.6 | 92.2 | 93.5 | 69.8 | 93.4 |

Where mAP is mean average precision; YOLO stands for You Only Look Once, ALPD stands for Auto-labeling of large poultry datasets, RT-DETR stands for Real-Time DEtector Transformer; Supervised models are trained on 400 images of each broiler and hen; Semi-supervised models are trained on 50 images of each broiler and hen.

The semi-supervised models, particularly those incorporating ALPD, also showed promising results. YOLOv8s-ALPD achieved an overall mAP_0.50 of 96.3%, with broiler precision and recall at 97.2% and 96.9%, respectively. YOLOv9s-ALPD further improved upon this with an overall mAP_0.50 of 95.7%, maintaining competitive performance for broiler detection while outperforming YOLOv8s-ALPD for hen detection with a recall of 87.3% and mAP_0.50 of 92.9%.

These findings are consistent with recent studies, where integrating semi-supervised learning approaches has enhanced model robustness, particularly in cases with limited labeled data (Kim et al., 2022; Shen et al., 2024). The ALPD variants, such as YOLO-World-ALPD, also maintained high detection capabilities, with broiler performance like their supervised counterparts. Notably, semi-supervised models exhibited a trade-off in performance, with slight reductions in overall precision but gains in flexibility when data availability is limited. This aligns with previous research by Zhang et al. (2022) and Liu et al. (2022), who observed that while supervised models tend to perform better in accuracy, semi-supervised learning offers the advantage of improving model performance in real-world applications with minimal labeled data.

The training dataset outputs depicted in **Fig. 5** reveal notable distinctions in loss performance across models (box_loss, cls_loss, and dfl_loss). Box loss measures bounding box accuracy, cls loss evaluates classification correctness, and dfl loss refines box localization precision (Paneru et al., 2024). YOLOv8s variants, particularly YOLOv8s-World-ALPD, exhibit lower training losses than other models, signifying efficient optimization and effective handling of training data. Studies have shown that reducing training or validation losses are correlated with higher detection accuracy in deep-learning models (H. Li et al., 2024; Munir et al., 2023). RT-DETR achieves exceptionally low training losses, particularly in box_loss, reflecting its strong performance during training and potential for robust predictions. In contrast, Faster R-CNN consistently achieves the lowest training box_loss and cls_loss among all models; however, its performance does not translate effectively to validation datasets, indicating a possible overfitting issue. This happens when the model learns the training data too well, including noise and outliers, which reduces its ability to generalize to new, unseen data (Jabbar & Khan, 2015). In addition, the Faster R-CNN model found struggling with small or unusually shaped objects, heavily occluded items, and cluttered scenes (Boesch, 2024). Models like YOLOv10s and YOLOv11s display higher training losses, suggesting suboptimal convergence or limitations in architecture refinement during training. Including ALPD markedly improves training efficiency across YOLO models, reducing losses and indicating enhanced label quality and dataset scalability. These results underline the superior training performance of YOLOv8s variants and RT-DETR, while highlighting the limitations of Faster R-CNN despite its low training losses. This suggests the need for more effective regularization or validation strategies to prevent overfitting in its application.

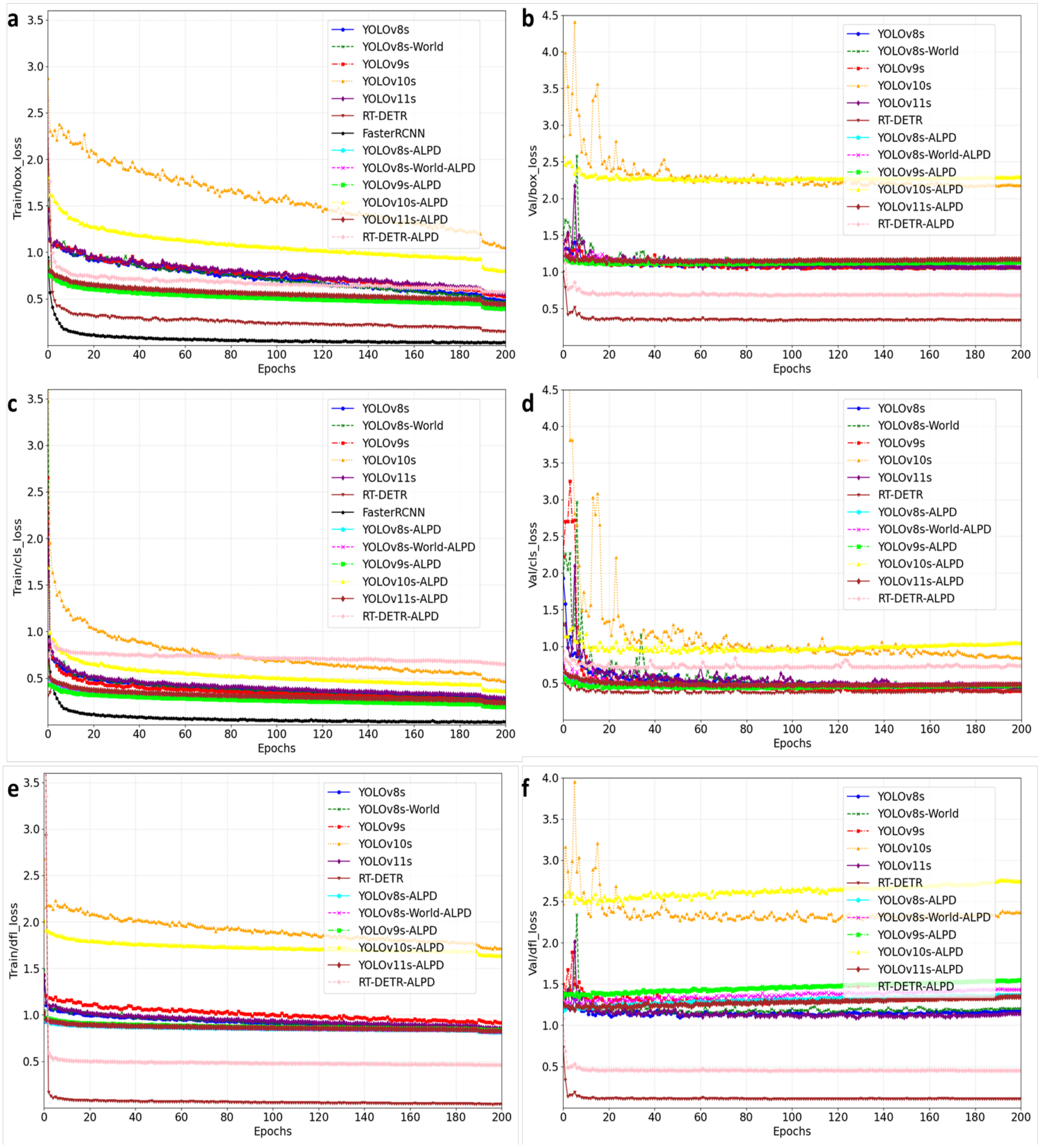

**Fig. 5.** Performance comparison of different models for broiler and hen detection across a) Train box loss, b) Validation box loss, c) Train class loss, d) Validation class loss, e) Train dfl loss, and f) Validation dfl loss. YOLO is You Only Look Once; ALPD is Auto-labeling of large poultry

datasets; RT-DETR is Real-Time DEtector TRansformer; Val is validation; cls is class; dfl is distribution focal loss.

Although models often exhibit high accuracy on training and validation datasets, their performance on test datasets can sometimes be less consistent, demonstrating a gap between training success and real-world applicability (Xu & Goodacre, 2018). This highlights the necessity of evaluating predicted outcomes on test data, as it provides insight into the model's ability to generalize beyond the data it was trained on. Such evaluations are critical for detecting issues like overfitting and ensuring the model's robustness in diverse, unseen scenarios (Jabbar & Khan, 2015). Consequently, thorough testing and performance assessment on test data are essential for validating a model's true predictive power and reliability. Therefore, the predicted performance of various models for detecting broilers and hens was evaluated across three training paradigms: zero-shot, supervised, and semi-supervised learning (**Table 5**). Zero-shot models like GDINO and YOLOWorld showed limited success in distinguishing between broilers and hens. While GDINO achieved moderate performance for broilers (MAE of 0.88 and F1 score of 79%), but it struggled to detect hens accurately, with a false-negative rate (FNR) of 100% and undefined precision and recall. During our analysis, we observed the presence of outliers and noise in the detected instances, which were mitigated using a normal distribution filter constrained within Mean ± 2 Standard Deviations. Additionally, overlapping bounding boxes, which can lead to ambiguity in object localization, were corrected by adjusting the non-maximum suppression value. This optimization ensured that the final detections were distinct and representative of true object boundaries. This approach effectively filtered out erroneous detections and enhanced the clarity of the resulting images, ensuring higher-quality inputs for downstream tasks. These results highlight Grounding DINO's strong detection capabilities, particularly when combined with statistical filtering techniques, and suggest its potential for broad applications in tasks requiring reliable and adaptive detection mechanisms. Similarly, YOLOWorld and GDINO exhibited high FNRs for hens, highlighting their limitations in multi-class detection. Notably, CLIP performed overall better with a precision of 76.5%, recall of 64.1%, and an F1 score of 69.65. However, the zero-shot models lacked sufficient generalizability for differentiating between broilers and hens, primarily due to fine-tuning and lack of training on similarly labeled datasets.

**Table 5.** Predicted performance metrics of zero-shot, supervised, and semi-supervised models for broiler and hen detection across various models.

| Categories | Models | Class | MAE | MSE | RMSE | Precision (%) | Recall (%) | F1 Score (%) | FNR (%) |
|---|---|---|---|---|---|---|---|---|---|
| Zero-shot* | GDINO | Broiler | 0.9 | 213.0 | 14.6 | 94.7 | 68.0 | 79.0 | 32.0 |
| | | Hen | 1.2 | 368.0 | 19.2 | 0.0 | 0.0 | Undefined | 100.0 |
| | | Overall | 1.1 | 290.5 | 16.9 | 47.4 | 34.0 | 79.0 | 66.0 |
| | YOLOWorld | Broiler | 22.8 | 700.0 | 26.5 | 50.1 | 50.2 | 50.2 | 49.8 |
| | | Hen | 17.0 | 742.0 | 27.2 | 0.0 | 0.0 | Undefined | 100.0 |
| | | Overall | 19.9 | 721.0 | 26.9 | 25.1 | 25.1 | 50.2 | 74.9 |
| | CLIP | Broiler | 1.1 | 163.0 | 12.8 | 36.5 | 24.9 | 29.7 | 75.1 |
| | | Hen | 1.0 | 281.0 | 16.7 | 100.0 | 96.7 | 98.0 | 3.3 |
| | | Overall | 1.1 | 222.0 | 14.8 | 68.3 | 60.8 | 63.9 | 39.2 |
| | YOLOv8s | Broiler | 0.4 | 0.8 | 0.9 | 98.9 | 99.5 | 99.2 | 0.5 |
| | | Hen | 0.9 | 3.2 | 1.8 | 95.2 | 99.9 | 97.5 | 0.1 |
| | | Overall | 0.7 | 2.0 | 1.4 | 97.1 | 99.7 | 98.4 | 0.3 |
| | | Broiler | 0.3 | 0.4 | 0.6 | 99.2 | 99.5 | 99.4 | 0.5 |
| | **YOLOv8s-World** | **Hen** | 0.6 | 1.9 | 1.4 | 97.1 | 99.2 | 98.1 | 0.8 |
| | | **Overall** | 0.5 | 1.2 | 1.0 | 98.2 | 99.4 | 98.8 | 0.6 |
| **Supervised** | **YOLOv9s** | **Broiler** | 0.4 | 0.6 | 0.8 | 99.1 | 99.4 | 99.2 | 0.6 |
| | | **Hen** | 0.5 | 1.6 | 1.3 | 97.3 | 99.9 | 98.6 | 0.1 |
| | | **Overall** | 0.5 | 1.1 | 1.1 | 98.2 | 99.7 | 98.9 | 0.3 |
| | **YOLOv10s** | **Broiler** | 0.9 | 2.7 | 1.7 | 96.8 | 99.5 | 98.1 | 0.5 |
| | | **Hen** | 1.1 | 3.7 | 1.9 | 95.4 | 98.5 | 96.9 | 1.5 |
| | | **Overall** | 1.0 | 3.2 | 1.8 | 96.1 | 99.0 | 97.5 | 1.0 |
| | **YOLOv11s** | **Broiler** | 0.4 | 0.9 | 1.0 | 98.6 | 99.6 | 99.1 | 0.4 |
| | | **Hen** | 1.0 | 6.1 | 2.5 | 94.7 | 99.7 | 97.2 | 0.3 |
| | | **Overall** | 0.7 | 3.5 | 1.8 | 96.7 | 99.7 | 98.2 | 0.3 |
| | **RT-DETR-l** | **Broiler** | 0.5 | 1.9 | 1.4 | 96.4 | 98.3 | 99.1 | 1.7 |
| | | **Hen** | 1.0 | 2.9 | 1.7 | 95.2 | 98.1 | 97.4 | 1.9 |
| | | **Overall** | 0.6 | 2.4 | 1.6 | 95.8 | 98.2 | 98.3 | 1.8 |
| | **Faster-RCNN** | **Broiler** | 1.1 | 4.7 | 2.2 | 90.0 | 96.5 | 93.1 | 3.5 |
| | | **Hen** | 2.4 | 7.5 | 2.7 | 87.8 | 93.8 | 90.7 | 6.2 |
| | | **Overall** | 1.8 | 6.1 | 2.5 | 88.9 | 95.2 | 91.9 | 4.8 |
| **Semi-supervised** | **YOLOv8s-ALPD** | **Broiler** | 1.0 | 3.4 | 1.8 | 96.7 | 99.2 | 97.9 | 0.8 |
| | | **Hen** | 1.0 | 3.6 | 1.9 | 95.6 | 98.9 | 97.2 | 1.1 |
| | | **Overall** | 1.0 | 3.5 | 1.9 | 96.2 | 99.1 | 97.6 | 0.9 |
| | **YOLO-World-ALPD** | **Broiler** | 0.9 | 2.8 | 1.7 | 96.5 | 99.7 | 98.1 | 0.3 |
| | | **Hen** | 1.8 | 11.7 | 3.4 | 90.8 | 99.9 | 95.1 | 0.1 |
| | | **Overall** | 1.4 | 7.3 | 2.6 | 93.7 | 99.8 | 96.6 | 0.2 |

| | | | | | | | | |
|---|---|---|---|---|---|---|---|---|
| **YOLOv9s-ALPD** | **Broiler** | 0.8 | 2.6 | 1.6 | 97.1 | 99.6 | 98.3 | 0.4 |
| | **Hen** | 1.3 | 6.8 | 2.6 | 93.4 | 99.3 | 96.3 | 0.7 |
| | **Overall** | 1.1 | 4.7 | 2.1 | 95.3 | 99.5 | 97.3 | 0.5 |
| **YOLOv10s-ALPD** | **Broiler** | 1.0 | 3.5 | 1.9 | 96.3 | 99.3 | 97.8 | 0.7 |
| | **Hen** | 1.2 | 4.4 | 2.1 | 93.9 | 99.3 | 96.5 | 0.7 |
| | **Overall** | 1.1 | 4.0 | 2.0 | 95.1 | 99.3 | 97.2 | 0.7 |
| **YOLOv11s-ALPD** | **Broiler** | 0.9 | 3.5 | 1.9 | 96.4 | 99.7 | 98.0 | 0.3 |
| | **Hen** | 2.1 | 14.3 | 3.8 | 89.1 | 99.8 | 94.2 | 0.2 |
| | **Overall** | 1.5 | 8.9 | 2.9 | 92.8 | 99.8 | 96.1 | 0.2 |

*Model performance is higher when just detecting one class, either as "broiler, chicken, or bird," but when it comes to detecting between two classes, "broiler" and "hen," neither of the models can detect hen; instead, it is mislabeled as "broiler" or "chicken." Supervised models are trained on 400 images of each broiler and hen; Semi-supervised models are trained on 50 images of each broiler and hen; YOLO is You Only Look Once; CLIP is Contrastive Language–Image Pretraining; GDINO is Ground DINO; ALPD is Auto-labeling of large poultry datasets; MAE is the mean absolute error; MSE is the mean squared error; RMSE is the root mean squared error; FNR is the false negative rate.

Semi-supervised models demonstrated remarkable performance despite being trained on a significantly smaller dataset of just 50 labeled images (30 for training, 10 for validation, and 10 for testing), as shown in **Fig. 6.** For instance, YOLOv8s-ALPD achieved an overall F1 score of 97.57% and a low FNR of 0.96%, closely matching the results of fully supervised models trained on 400 labeled images (240 for training, 80 for validation, 80 for testing). The semi-supervised YOLOv9s-ALPD and YOLO-World-ALPD models also maintained competitive accuracy, achieving F1 scores of 97.29% and 96.57%, respectively. In contrast, supervised models like YOLOv8s and YOLOv8s-World achieved slightly higher F1 scores (98.34% and 98.74%) but required substantially more labeled data. These results underscore the efficiency of semi-supervised approaches, which achieved near-supervised performance while reducing the dependency on large labeled datasets. This makes semi-supervised learning particularly valuable when data labeling is costly or time-consuming.

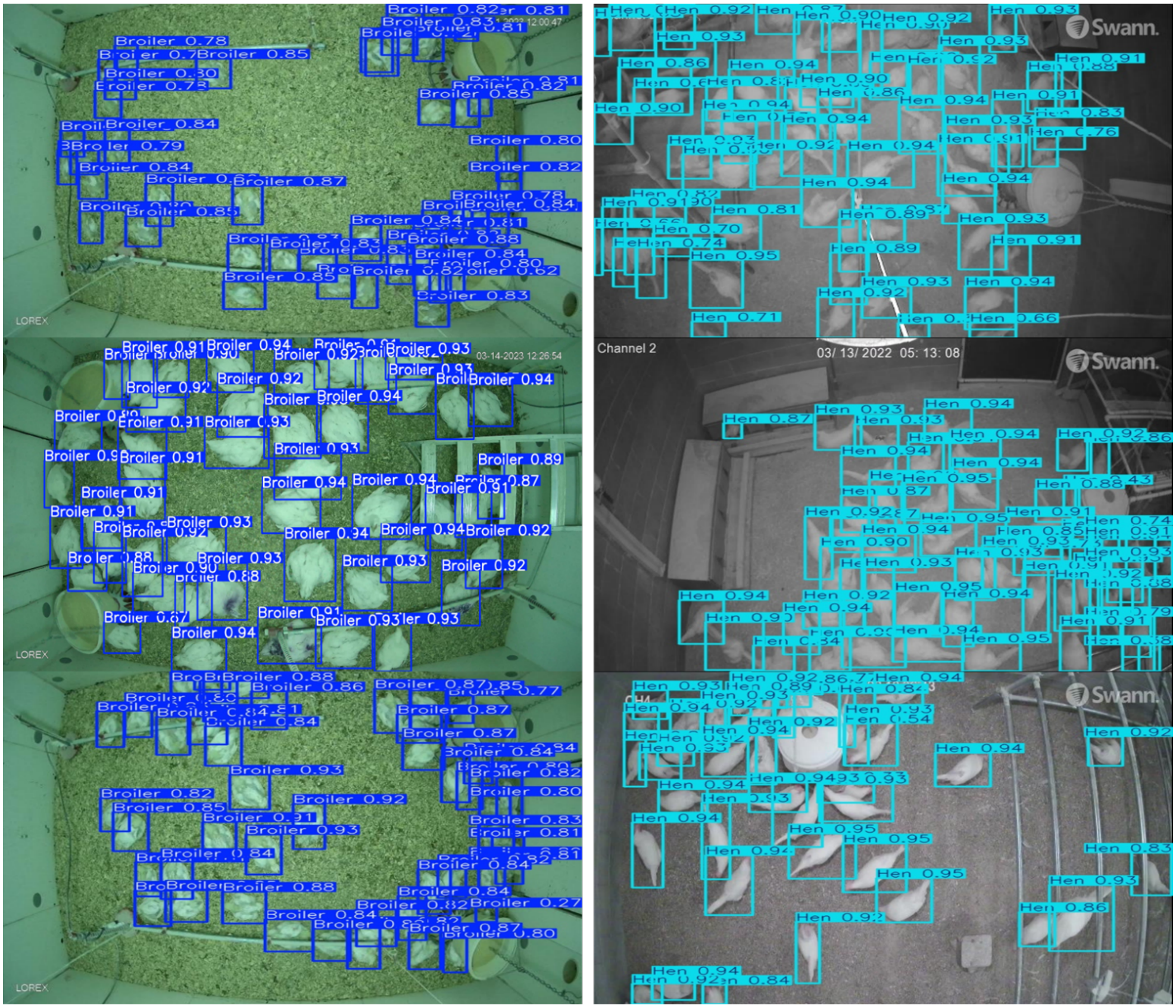

**Fig. 6.** Detection results of broiler (right) and hen (left) using different models.

*3.3. Hybrid AI detection model performance evaluation using text-based embeddings*

**Table 6** presents the performance metrics of various hybrid detection models for broilers and hens. Among the models, YOLO + CLIP and YOLOWorld + CLIP demonstrated competitive results, showcasing their ability to combine strong feature extraction with detection. The YOLOWorld + CLIP hybrid achieved an overall F1 score of 97.55%, with a precision of 99.1% and recall of 96.1%. This model exhibited superior performance for broilers, with a mean absolute error (MAE) of 0.31 and a root mean squared error (RMSE) of 0.64, highlighting its robustness in identifying broilers with minimal false negatives (FNR of 0.656). However, the model's performance declined slightly for hens, evidenced by a higher FNR of 7.15 and an RMSE of 2.57, suggesting room for improvement in hen detection. Comparatively, YOLO-World with the YOLOv8 backbone outperformed other hybrids' overall detection accuracy, achieving an F1 score of 98.74%, with a balanced precision (98.15%) and recall (99.35%). This model demonstrated excellent detection of both broilers and hens (**Fig. 7**), with notably low MAE (0.3 for broilers and 0.64 for hens), underscoring its robustness and consistency.

**Table 6.** Predicted performance metrics comparison of hybrid detection models.

| Hybrid detection Model | Class | MAE | MSE | RMSE | Precision | Recall | F1 Score | FNR |
|---|---|---|---|---|---|---|---|---|
| **YOLO + CLIP** | **Broiler** | 0.8 | 2.5 | 1.6 | 97.2 | 99.5 | 98.3 | 0.6 |
| | **Hen** | 1.4 | 5.9 | 2.4 | 97.6 | 94.3 | 96.0 | 5.7 |
| | **Overall** | 1.1 | 4.2 | 2.0 | 97.4 | 96.9 | 97.2 | 3.2 |
| **YOLOWorld +CLIP** | **Broiler** | 0.3 | 0.4 | 0.6 | 99.3 | 99.3 | 99.3 | 0.7 |
| | **Hen** | 1.4 | 6.6 | 2.6 | 98.9 | 92.9 | 95.8 | 7.2 |
| | **Overall** | 0.9 | 3.5 | 1.6 | 99.1 | 96.1 | 97.6 | 4.0 |
| **YOLO-World + YOLOv8** | **Broiler** | 0.3 | 0.4 | 0.6 | 99.2 | 99.5 | 99.4 | 0.5 |
| | **Hen** | 0.6 | 1.9 | 1.4 | 97.1 | 99.2 | 98.1 | 0.8 |
| | **Overall** | 0.5 | 1.2 | 1.09 | 98.2 | 99.4 | 98.8 | 0.7 |
| **YOLO + GDINO** | **Broiler** | 11.4 | 366.0 | 19.1 | 99.7 | 50.2 | 66.8 | 49.8 |
| | **Hen** | 5.6 | 104.0 | 10.2 | 99.9 | 67.1 | 80.3 | 32.9 |
| | **Overall** | 8.5 | 235.0 | 14.7 | 99.8 | 58.7 | 73.6 | 41.4 |

Where YOLO is You Only Look Once; CLIP is Contrastive Language–Image Pretraining; GDINO is Ground DINO; MAE is the mean absolute error; MSE is the mean squared error; RMSE is the root mean squared error; FNR is the false negative rate.

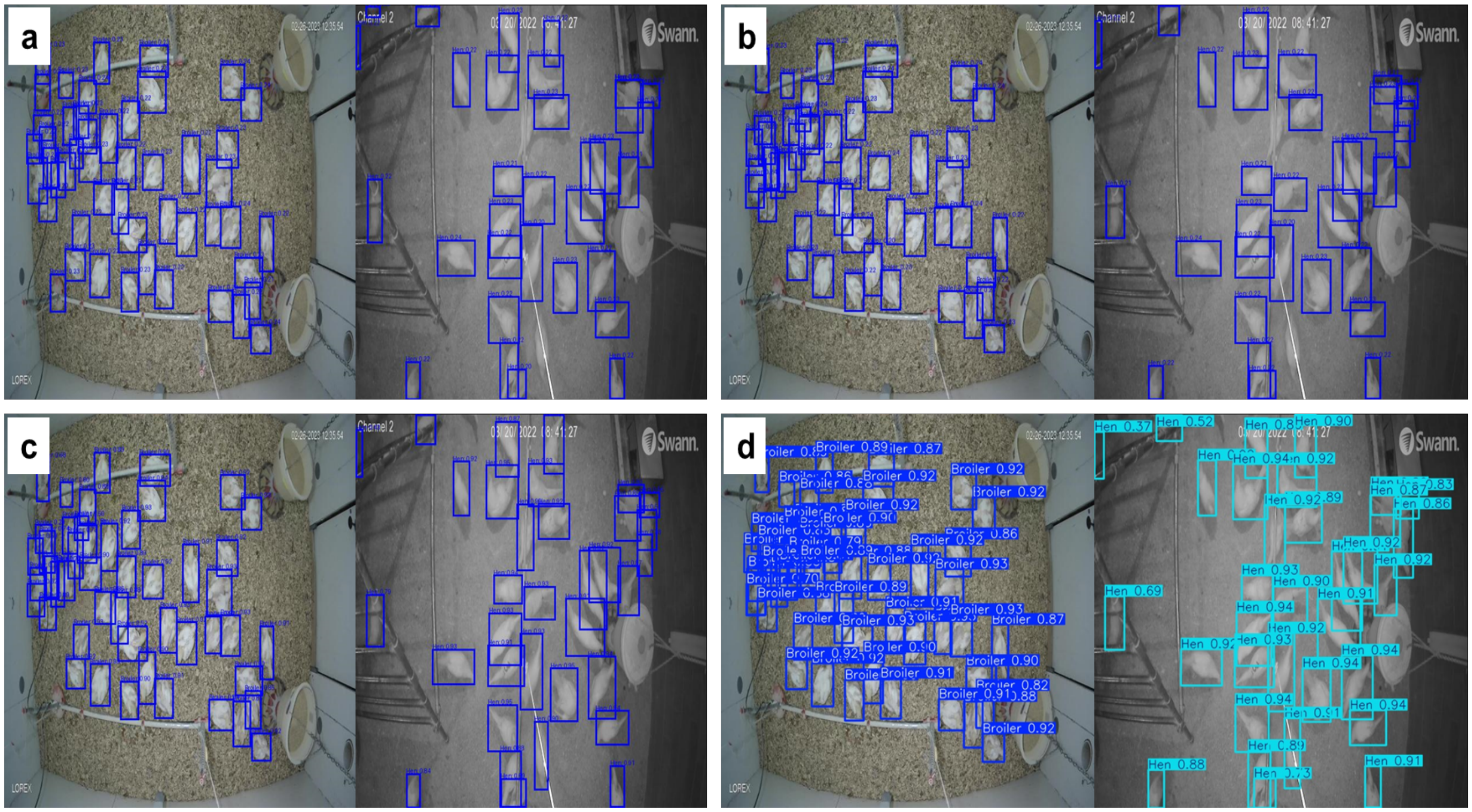

**Fig. 7.** Detected results for broilers and hens using hybrid models a) YOLO + CLIP, b) YOLOWorld + CLIP, c) YOLO-Worldv2 + YOLOv8 backbone, and d) YOLO + GDINO. The left

side shows detected broilers, while the right side shows detected hens. YOLO is You Only Look Once; CLIP is Contrastive Language–Image Pretraining; GDINO is Ground DINO.

Similarly, the YOLO + GDINO hybrid, combining a zero-shot model with a supervised detection model, showed unique performance trends. While the overall F1 score (73.54%) and recall (58.66%) were lower compared to other hybrids, this model exhibited high precision (99.78%), particularly excelling in broiler detection (precision of 99.67%). These results highlight the potential of zero-shot models when integrated with supervised architectures, as they may enhance certain performance aspects like precision while compensating for deficiencies in recall. However, the YOLO + GDINO model struggled with higher error metrics, such as MAE (8.51) and RMSE (14.7), which indicate reduced generalization across classes. These findings suggest that while zero-shot models alone may not be optimal, their strategic integration with robust supervised and semi-supervised models can unlock new avenues for performance enhancement in hybrid systems.

*3.4. Different amounts of image labeling*

**Table 7** and **Fig. 8** illustrate the performance metrics for broiler and hen detection across varying image quantities (I50, I100, I200, and I400). The results reveal a clear trend in the enhancement of model performance with the increase in the number of training images. For broiler detection, all key metrics, including precision, recall, F1-score, mAP_0.50, and mAP_0.50-0.95, consistently improved as the dataset size increased. At I50, the precision was 94%, and recall was 96.6%, progressively increasing to 98.1% and 98.6%, respectively, with 400 images. The mAP_0.50 also showed a steady rise from 97.8% at I50 to 99.3% at I400, with the F1-score following a similar trajectory from 95.3% to 98.3%. This trend indicates that increasing the dataset size enhances the model's ability to detect broilers more accurately and consistently, improving generalization and reducing overfitting. These results are consistent with the theory that larger training datasets enable the model to capture more diverse variations in the target object, thereby improving detection performance (Djolonga et al., 2021).

**Table 7.** Validation performance metrics for broiler and hen detection across various image quantities.

| Name | Class | Precision | Recall | mAP_0.50 | mAP_0.50-0.95 | F1-score |
|---|---|---|---|---|---|---|
| | **Broiler** | 94.0 | 96.6 | 97.8 | 69.6 | 95.3 |
| **I50** | **Hen** | 93.7 | 82.1 | 92.9 | 54.2 | 87.5 |
| | **Overall** | 93.8 | 89.3 | 95.4 | 61.9 | 91.5 |
| | **Broiler** | 92.3 | 96.5 | 97.9 | 71.6 | 94.4 |
| **I100** | **Hen** | 95.9 | 75.0 | 94.7 | 61.9 | 84.2 |
| | **Overall** | 94.1 | 85.8 | 96.3 | 66.7 | 89.8 |
| | **Broiler** | 95.2 | 97.1 | 98.8 | 72.6 | 96.1 |
| **I200** | **Hen** | 94.8 | 85.3 | 95.2 | 66.6 | 89.8 |
| | **Overall** | 95.0 | 91.2 | 97.0 | 69.6 | 93.1 |
| | **Broiler** | 98.1 | 98.6 | 99.3 | 72.9 | 98.3 |
| **I400** | **Hen** | 94.9 | 91.0 | 96.3 | 69.0 | 92.9 |
| | **Overall** | 96.5 | 94.8 | 97.8 | 71.0 | 95.6 |

where mAP is the mean average precision; I50, I100, I200, and I400 are the quantities 50, 100, 200, and 400 images, respectively.

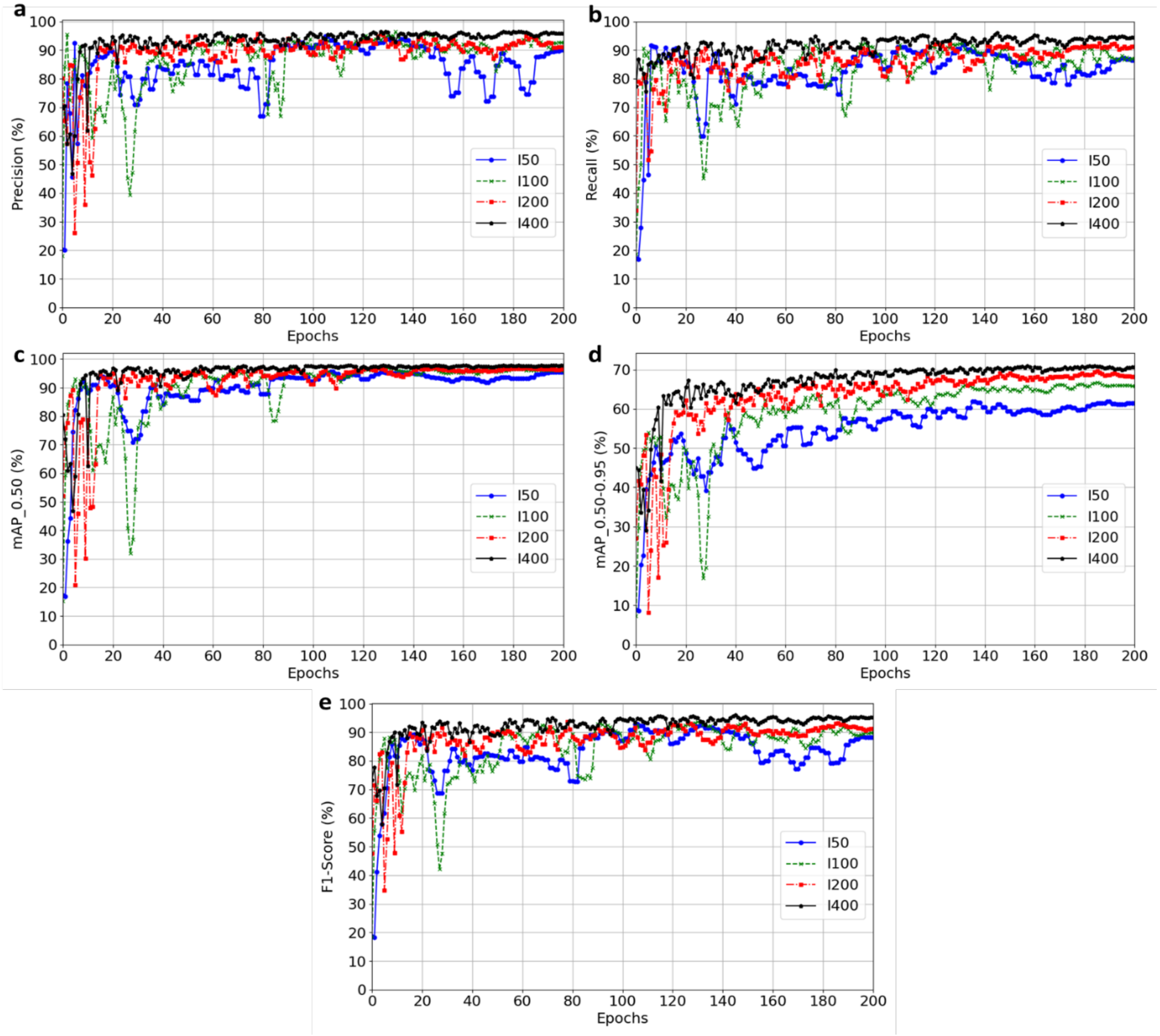

**Fig. 8.** Performance comparison of different image quantities for broiler and hen detection with a) precision, b) recall, c) mAP_0.50, d) mAP_0.50-0.95, and e) F1-score. mAP is the mean average precision; I50, I100, I200, and I400 are the quantities 50, 100, 200, and 400 images, respectively.

In contrast, while showing improvements across all metrics, hen detection did not exhibit as substantial gains as broiler detection with the increasing dataset size. At I50, precision was 93.7%, and recall was 82.1%, which gradually improved to 94.9% and 91% at I400. The mAP_0.50 metric also showed a similar improvement, rising from 92.9% with I50 to 96.3% with I400, while the F1-score increased from 87.5% to 92.9%. These findings suggest that although expanding the dataset improves hen detection performance, the model’s ability to capture the unique characteristics of hens was somewhat limited compared to broilers. Nonetheless, the consistent upward trend across all metrics confirms that more training data, even if less influential than for broiler detection, contributes positively to the model's accuracy and robustness. Overall, the data demonstrates that increasing the number of images for training significantly enhances the performance of detection models, particularly for broilers, and although the improvements for hens were relatively less pronounced. This still reflects the benefits of incorporating more diverse data into the training process. However, large datasets may not always be beneficial; proper regularization and handling of noisy examples or outliers are crucial (Zhu et al., 2012).

The performance comparison of the I50, I100, I200, and I400 datasets demonstrates that increasing the number of labeled images significantly enhances model performance, particularly in training (**Fig. 9**). I400, which uses the most labeled images, consistently shows the lowest training box_loss and classification loss, outperforming the other models. While I400 exhibits the highest training dfl_loss, this trade-off is acceptable, given the improvements in classification accuracy. In validation performance, I400 maintains competitive results with a box_loss and a classification loss, reflecting a strong ability to generalize despite the increased complexity of the dataset. This suggests that incorporating more labeled images improves training accuracy and enhances the model's ability to handle diverse data. It leads to better overall performance. Therefore, I400 demonstrates the benefits of using larger labeled datasets, resulting in improved model accuracy and robustness.

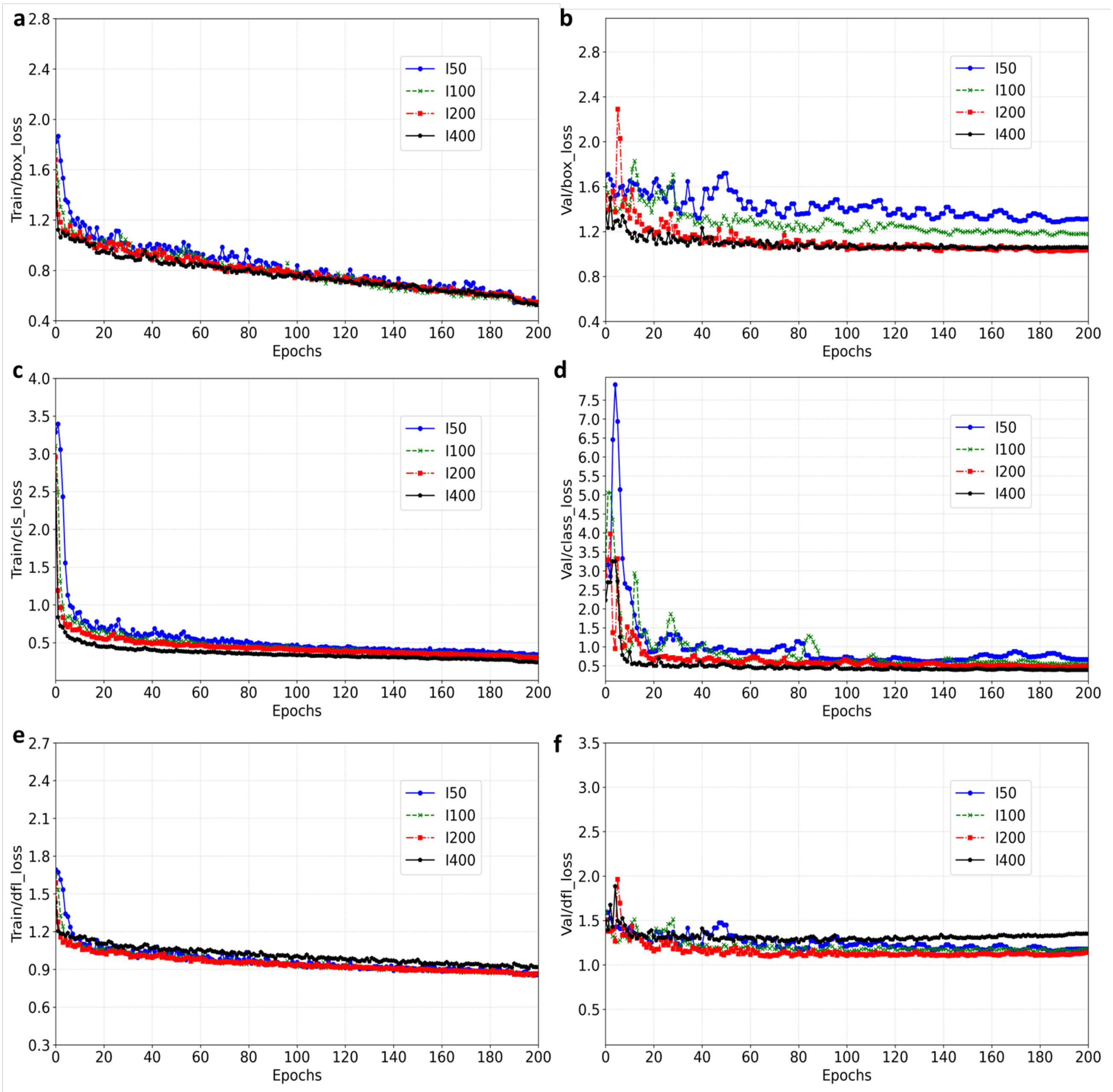

**Fig. 9.** Performance comparison of varying image quantity for broiler and hen detection across a) Train box loss, b) Validation box loss, c) Train class loss, d) Validation class loss, e) Train dfl loss, and f) Validation dfl loss. mAP is the mean average precision; Val is validation; cls is class; dfl is distribution focal loss; I50, I100, I200, and I400 are the quantities 50, 100, 200, and 400 images, respectively.

Similarly, the confusion matrices in **Fig. 10** demonstrate the detection performance of models trained on datasets with increasing image quantities: I50, I100, I200, and I400. The results reveal an improvement in detection accuracy as the dataset size increases. Higher true positive detection generally leads to better accuracy in a model. This is because accuracy measures the proportion of correct predictions (both true positives and true negatives) out of all predictions made (Bist et al., 2023). For I50, the broiler detection confidence score accuracy is 98%, with minor misclassifications into other categories. However, the hen detection is less robust, with an accuracy of 84% and considerable misclassification as the background. As the dataset grows to I100 and I200, the detection performance for hens improves to 83% and 88%, respectively, with reduced misclassification. The I400 dataset yields the best performance, with broiler and hen detection accuracies of 99% and 93%, respectively, and minimal misclassification. These findings underscore the importance of larger datasets in enhancing detection accuracy, particularly for challenging classes such as hens, while also reducing false positives in the background class.

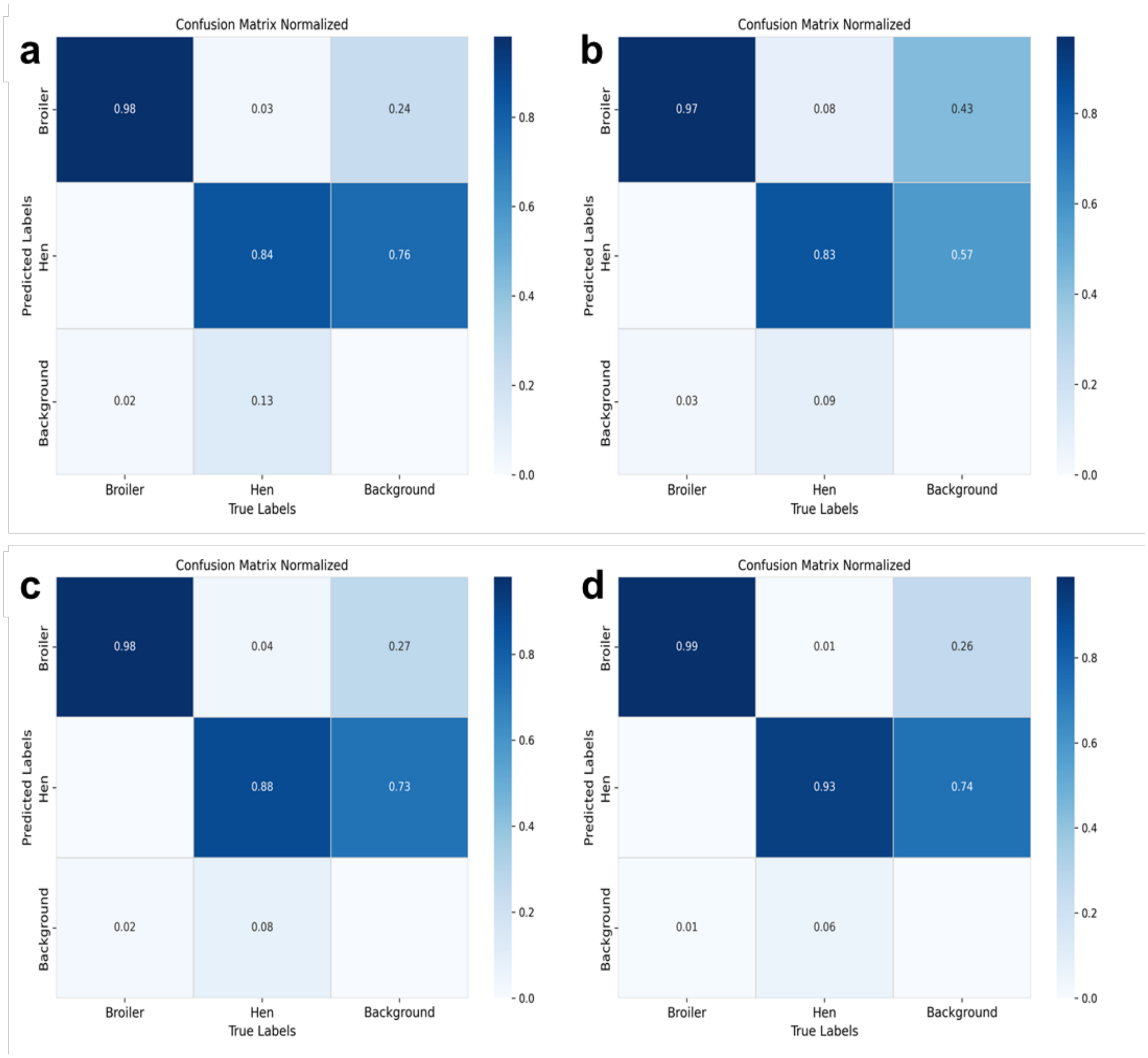

**Fig. 10.** Confusion matrices for hen and broiler detection using models trained on datasets with varying image quantities: a) I50, b) I100, c) I200, and d) I400. mAP is the mean average precision; I50, I100, I200, and I400 are the quantities 50, 100, 200, and 400 images, respectively.

**Table 8** presents the prediction performance metrics for broiler and hen detection across various image quantities. The results show a consistent improvement in model performance as the number of training images increases, particularly for broiler detection. At I50, the broiler detection model achieved a mean absolute error (MAE) of 0.85 and a root mean square error (RMSE) of 1.62, with a precision of 97%, recall of 99.3%, and an F1 score of 98.16%. As the dataset size increased, the MAE decreased to 0.35 and RMSE to 0.77 at I400, while precision and recall increased to 99.1% and 99.4%, respectively, with an F1 score of 99.24%. These results highlight the positive impact of larger datasets on broiler detection, with significant reductions in error and increase in precision and recall. The FNR also slightly increased to higher image quantities for broiler, particularly at I400. This indicates some trade-off between sensitivity and false negatives, but the model's overall performance improved substantially.

**Table 8.** Prediction performance metrics for broiler and hen detection across various image quantities.

| **Quantity** | **Class** | **MAE** | **MSE** | **RMSE** | **Precision** | **Recall** | **F1 Score** | **FNR (%)** |
|---|---|---|---|---|---|---|---|---|
| | **Broiler** | 0.9 | 2.6 | 1.6 | 97.0 | 99.3 | 98.2 | 0.7 |
| | **Hen** | 1.8 | 10.9 | 3.3 | 90.7 | 99.7 | 95.0 | 0.3 |
| **I50** | **Overall** | 1.4 | 6.8 | 2.5 | 93.9 | 99.5 | 96.6 | 0.5 |
| | **Broiler** | 0.9 | 3.5 | 1.9 | 96.5 | 99.4 | 98.0 | 0.4 |
| | **Hen** | 1.0 | 4.0 | 2.0 | 94.6 | 99.7 | 97.1 | 0.3 |
| **I100** | **Overall** | 1.0 | 3.8 | 2.0 | 95.6 | 99.6 | 97.6 | 0.4 |
| | **Broiler** | 0.4 | 0.9 | 1.0 | 98.6 | 99.5 | 99.1 | 0.6 |
| | **Hen** | 0.9 | 2.8 | 1.7 | 95.5 | 99.4 | 97.4 | 0.6 |
| **I200** | **Overall** | 0.7 | 1.9 | 1.4 | 97.1 | 99.5 | 98.3 | 0.6 |
| | **Broiler** | 0.4 | 0.6 | 0.8 | 99.1 | 99.4 | 99.2 | 0.6 |
| | **Hen** | 0.5 | 1.6 | 1.3 | 97.3 | 99.9 | 98.6 | 0.2 |
| **I400** | **Overall** | 0.5 | 1.1 | 1.1 | 98.2 | 99.7 | 98.9 | 0.4 |

where mAP is the mean average precision; MAE is the mean absolute error; MSE is the mean square error, RMSE is the root mean square error; FNR is the False negative rate; I50, I100, I200, and I400 are the quantities 50, 100, 200, and 400 images, respectively.

For hen detection, while the trend of improved performance with increasing image quantity was observed, the gains were less pronounced compared to broiler detection in predicted images (**Fig. 11**), the same as validation datasets. At I50, hen detection had a higher MAE of 1.78 and RMSE of 3.31, with a precision of 90.7% and recall of 99.7%. As the dataset size grew to I100 and I200, the MAE and RMSE decreased, with precision and recall improving, though the precision for hen detection at I200 remained somewhat lower compared to broiler detection. At I400, the MAE dropped to 0.49 and RMSE to 1.28, with precision reaching 97.3% and recall at 99.9%. The F1 score for hen detection also improved from 95.01% at I50 to 98.58% at I400, reflecting better model generalization. The false negative rate (FNR) was generally lower for hen detection, particularly at I400, where it decreased to 0.15%. These results demonstrate that while the model's performance for hen detection improved with more images, the relative improvements were less significant than for broilers. It can be possibly due to the more complex and varied characteristics of hens that were harder to capture with the dataset. Nevertheless, the overall findings indicate that increasing the image quantity enhances detection accuracy and robustness for both broilers and hens.

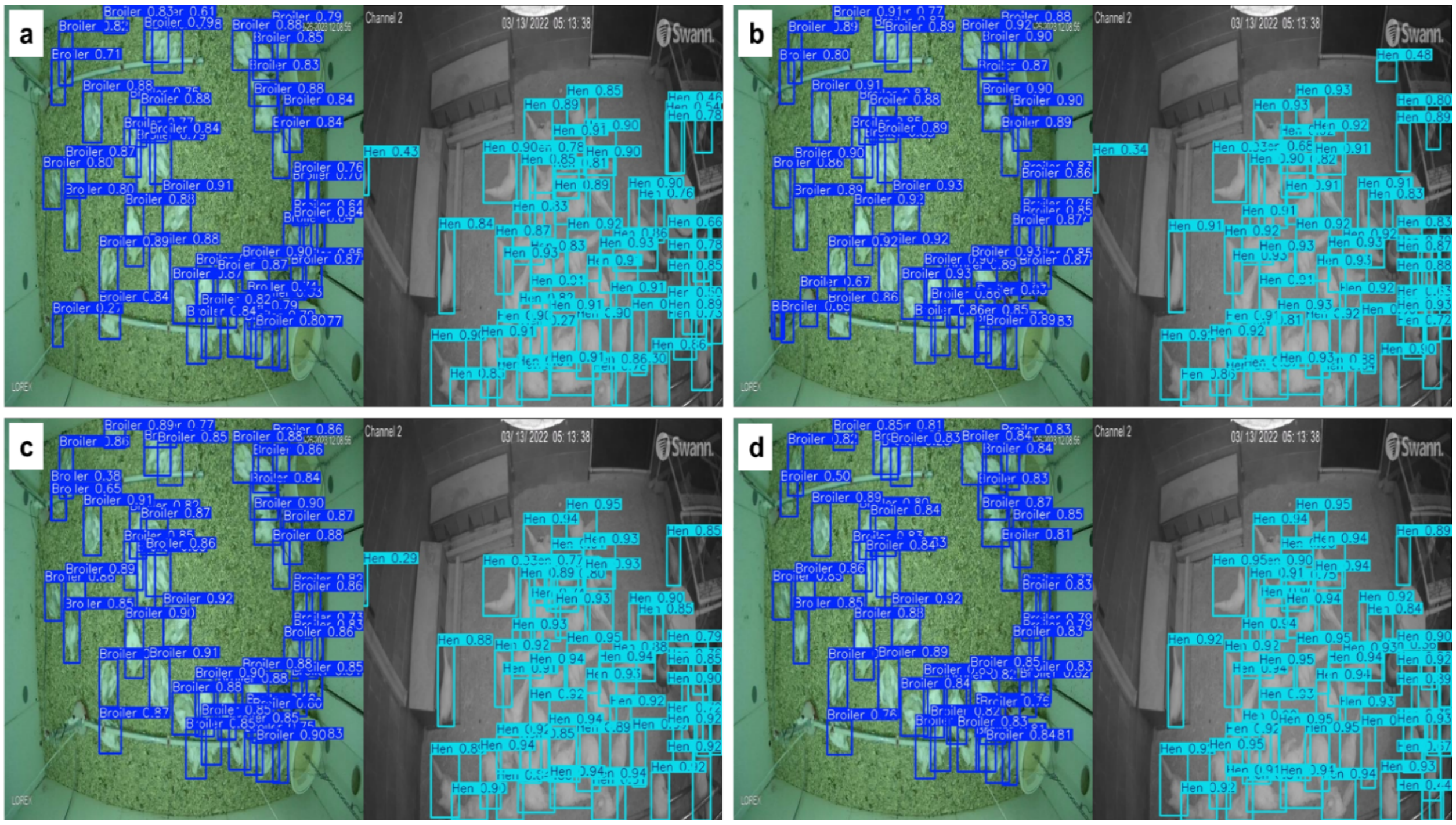

**Fig. 11.** Detected images of broilers and hens at different image quantities a) I50, b) I100, c) I200, and d) I400. The left side shows detected broilers, while the right side shows detected hens. I50, I100, I200, and I400 are the quantities 50, 100, 200, and 400 images, respectively.

*3.5. Comparison based on different confidence levels*

**Table 9** and **Fig. 12** present the performance metrics for broiler and hen detection across various confidence score thresholds. The results highlight the impact of confidence score levels on detection performance. At a 12.5% confidence level, broiler detection achieved a precision of 98.2%, recall of 97.8%, and an F1 score of 98.0%, while hen detection had a precision of 91.9%, recall of 92.8%, and an F1 score of 92.3%. The overall performance for both classes was strong, with the model achieving a precision of 95.0% and an F1 score of 95.1%. As the confidence threshold increased to 25%, 50%, and 75%, precision and recall for broiler detection showed slight improvements, with the highest performance at the 75% confidence level. Specifically, at 75%, broiler detection reached a precision of 98.7%, recall of 95.7%, and an F1 score of 97.2%, while hen detection reached a precision of 95.8%, recall of 88.3%, and an F1 score of 91.9%. Therefore, a higher confidence threshold leverages consistency regularization between strongly augmented views of an image using confidence-weighted pseudo-labels and demonstrates significant performance improvements over existing methods (Kim et al., 2022). However, this study observed the highest recall (94.5%), mAP_0.50 (97.8%), and F1-score (95.6%) at a 50% confidence interval. Higher confidence over 50% sometimes might miss true positive detections, reducing overall performance.

**Table 9.** Validation performance metrics for broiler and hen detection across various confidence scores.

| Confidence level | Class | Precision | Recall | mAP_ 0.50 | mAP_ 0.50-0.95 | F1-score |
|---|---|---|---|---|---|---|
| **Conf12.5** | **Broiler** | 98.2 | 97.8 | 99.1 | 73.2 | 98.0 |
| | **Hen** | 91.9 | 92.8 | 96.4 | 70.5 | 92.3 |
| | **all** | 95.0 | 95.3 | **97.8** | 71.8 | 95.1 |
| **Conf25** | **Broiler** | 98.7 | 97.6 | 99.1 | 73.6 | 98.1 |
| | **Hen** | 93.7 | 91.2 | 96.4 | 71.2 | 92.4 |
| | **all** | 96.2 | 94.4 | 97.8 | 72.4 | 95.3 |
| **Conf50** | **Broiler** | 98.4 | 97.7 | 99.1 | 74.0 | 98.1 |
| | **Hen** | 94.9 | 91.3 | 96.5 | 72.1 | 93.1 |
| | **all** | 96.7 | **94.5** | **97.8** | 73.0 | **95.6** |
| **Conf75** | **Broiler** | 98.7 | 95.7 | 98.8 | 74.9 | 97.2 |
| | **Hen** | 95.8 | 88.3 | 95.7 | 73.4 | 91.9 |
| | **all** | **97.2** | 92.0 | 97.3 | **74.2** | 94.5 |

Where, mAP- mean average precision; Conf12.5, Conf25, Conf50, and Conf75 are the confidence interval of 12.5%, 25%, 50%, and 75%, respectively.

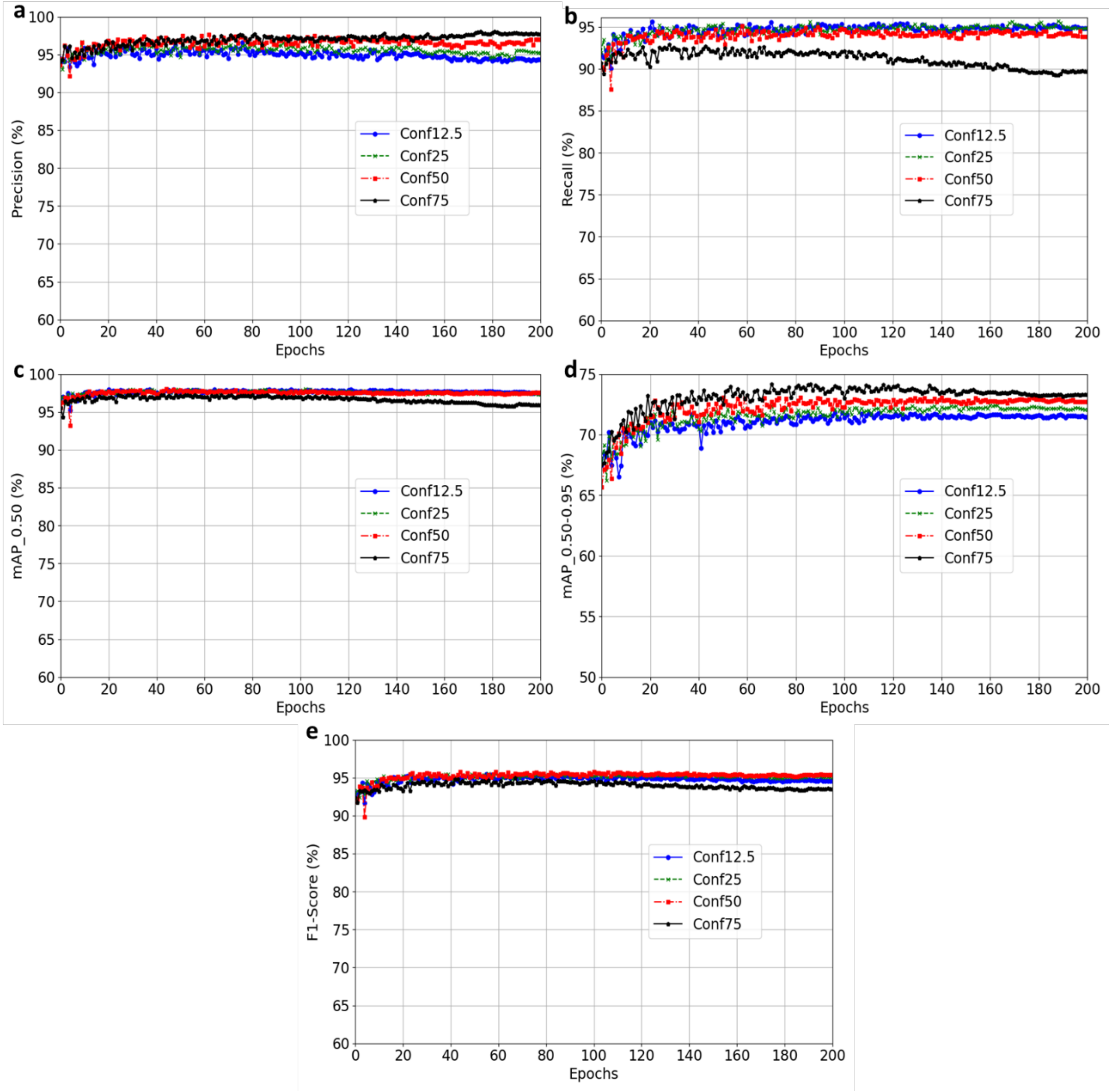

**Fig. 12.** Performance comparison of different confidence scores for broiler and hen detection with a) precision, b) recall, c) mAP_0.50, d) mAP_0.50-0.95, and e) F1-score. Where, mAP- means average precision; Conf12.5, Conf25, Conf50, and Conf75 are the confidence interval of 12.5%, 25%, 50%, and 75%, respectively.

Similarly, the overall results demonstrate that as the confidence threshold increased, there was a corresponding decrease in recall for hens. This suggests that higher thresholds may filter out more detections but also reduce the number of false positives. For broiler detection, however, higher confidence levels (especially at 50% and 75%) led to a more balanced improvement in precision and recall, with F1 scores improving from 98.0% at 12.5% to 97.2% at 75%. Despite these improvements, hen detection remained more challenging, with the highest F1 score of 93.1% observed at the 50% confidence threshold. This suggests that a mid-range confidence threshold (50%) may provide the best balance for hen detection, whereas broiler detection benefits from higher confidence levels, particularly at 75%. Overall, these findings suggest that confidence thresholds play a crucial role in optimizing detection performance. Lower thresholds are better for recall, but higher thresholds provide more precise results especially for broiler detection.

Moreover, the results from the training and validation datasets across different confidence intervals demonstrate how varying confidence thresholds affect model performance (**Fig. 13**). Training loss decreases as the confidence interval increases. Conf75 shows the lowest values for box_loss, cls_loss, and dfl_loss, indicating better optimization. Compared with Conf50, which shows slightly higher training losses (box_loss, cls_loss, and dfl_loss), Conf75 appears to be more efficient regarding training performance. However, Conf75 in the validation phase exhibits slightly higher val_box_loss and val_cls_loss than Conf50. It suggests that a higher confidence threshold improves training efficiency, however it may lead to a marginal decrease in generalization ability.

This suggests that increasing the confidence interval enhances training performance. It should be balanced against validation accuracy to ensure robust model performance across both phases.

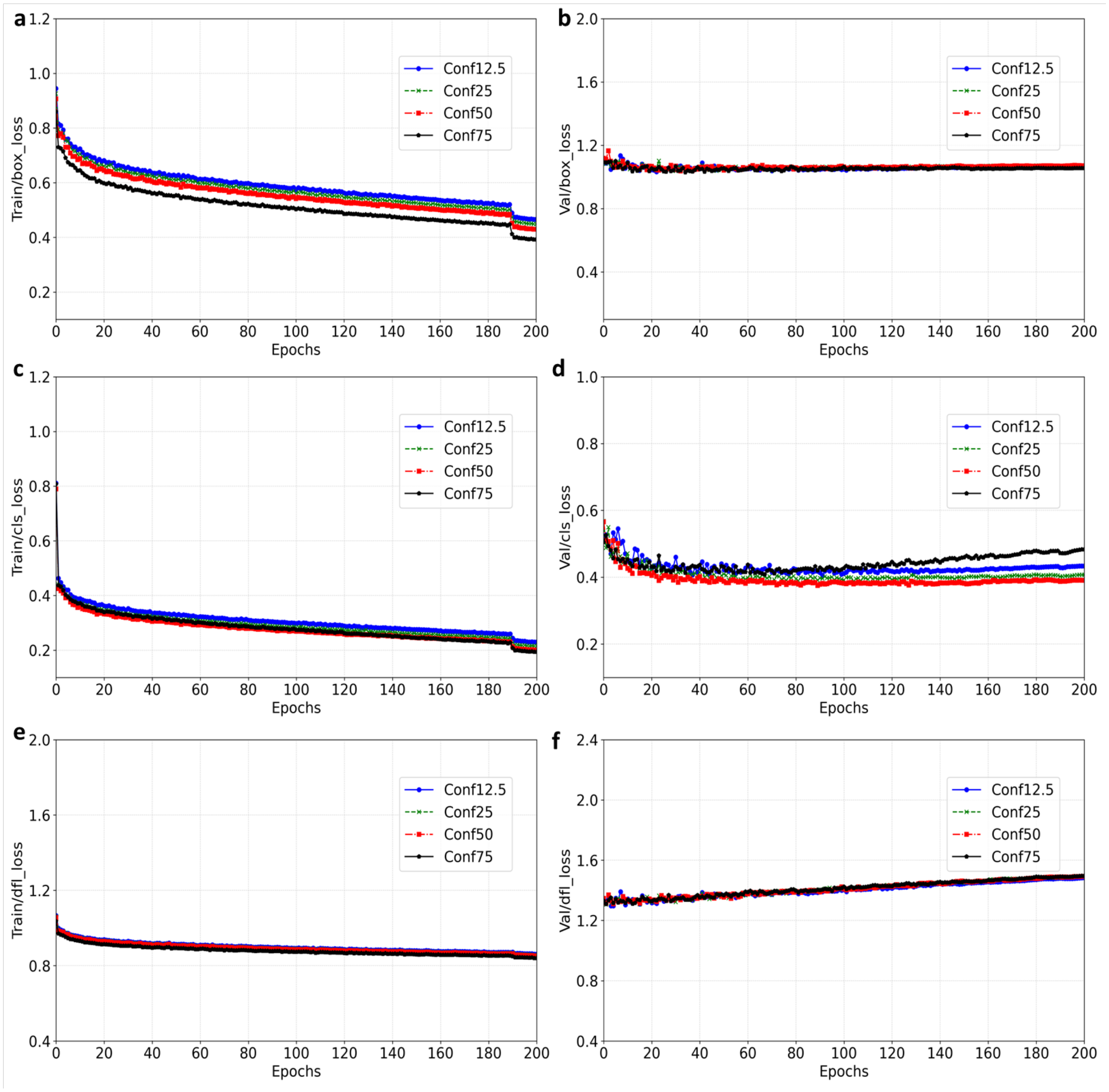

**Fig. 13.** Performance comparison based on varying confidence levels for broiler and hen detection across a) Train box loss, b) Validation box loss, c) Train class loss, d) Validation class loss, e) Train dfl loss, and f) Validation dfl loss. Where, mAP- mean average precision; Conf12.5, Conf25,

Conf50, and Conf75 are the confidence interval of 12.5%, 25%, 50%, and 75%, respectively; Val is validation; cls is class; dfl is distribution focal loss.

Similarly, the confusion matrices in **Fig. 14** illustrate the impact of varying confidence thresholds on the detection performance of broilers, hens, and backgrounds. At the lowest threshold (Conf12.5), the model achieves high detection counts for broilers and hens but with considerable misclassifications of hens into the background and false positives for the background. As the threshold increases to Conf25 and Conf50, the detection accuracy for hens improves, and background misclassifications decrease. However, broiler and hen detection accuracy at the highest threshold (Conf75) dropped significantly with broiler misclassification rising to 122 and background false positives increasing to 206. This analysis highlights a trade-off between confidence thresholds and detection performance. Higher thresholds reduce false positives but may compromise the overall detection accuracy for hens and broilers. Adjusting the confidence threshold optimally is critical to balancing precision and recall for effective detection in practical applications.

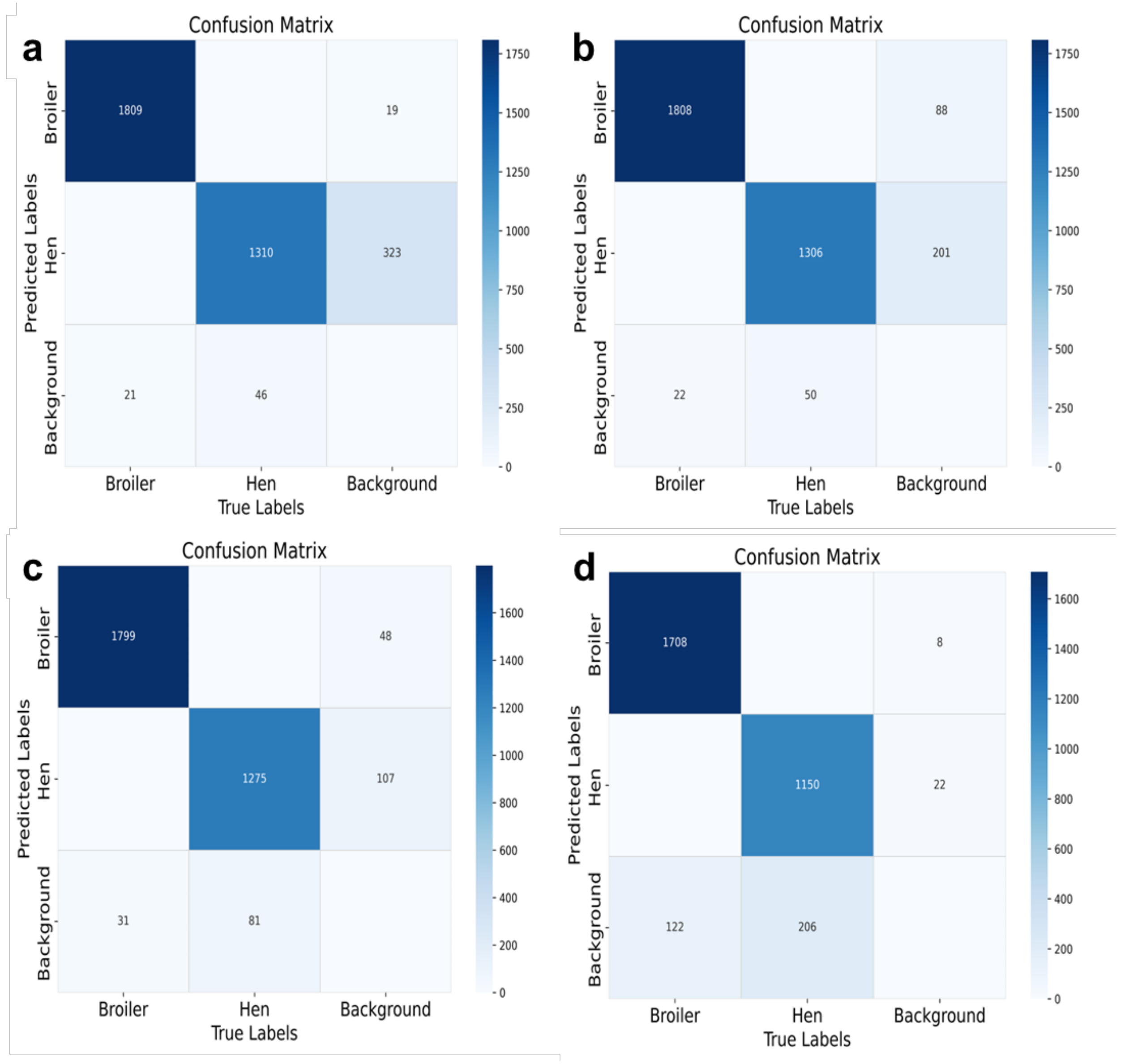

**Fig. 14.** Confusion matrices for hen and broiler detection across different confidence thresholds: a) Conf12.5, b) Conf25, c) Conf50, and d) Conf75. mAP- mean average precision; Conf12.5, Conf25, Conf50, and Conf75 are the confidence interval of 12.5%, 25%, 50%, and 75%, respectively.

Additionally, prediction results must be evaluated to determine the actual detection accuracy (**Fig. 15**). **Table 10** presents the prediction performance metrics for broiler and hen detection across different confidence levels. The model demonstrates high recall for broilers and hens at lower confidence levels (12.5% and 25%), with broiler detection achieving near-perfect recall at 99.67% and then detection achieving 100% recall at both levels. However, precision for hen detection remains lower compared to broilers, especially at 12.5%, where it drops significantly. As the confidence level increases, particularly at 50%, the model improves precision for both classes, with broiler detection achieving the highest precision and F1 score. Incorporating confidence thresholds ensures that only high-confidence pseudo-labeled samples are included in the retraining process, reducing errors and improving model reliability (Y. Wang et al., 2021). However, recall for broilers and hens at 50% shows a slight decrease compared to the lower thresholds, indicating a trade-off between recall and precision. The F1 score for broilers and hens peaks at this threshold, with the model effectively balancing both metrics.

**Table 10.** Prediction performance metrics for broiler and hen detection across various confidence levels.

| **Confidence level** | **Class** | **MAE** | **MSE** | **RMSE** | **Precision (%)** | **Recall (%)** | **F1 Score (%)** | **FNR (%)** |
|---|---|---|---|---|---|---|---|---|
| | **Broiler** | 1.4 | 7.4 | 2.7 | 94.6 | 99.7 | 97.1 | 0.3 |
| **Conf12.5** | **Hen** | 3.1 | 28.6 | 5.4 | 84.5 | 100.0 | 91.6 | 0.0 |
| | **Overall** | 2.3 | 18.0 | 4.1 | 89.6 | 99.9 | 94.4 | 0.2 |
| | **Broiler** | 0.9 | 3.1 | 1.8 | 96.4 | 99.7 | 98.0 | 0.3 |
| **Conf25** | **Hen** | 2.1 | 14.9 | 3.9 | 89.1 | 100.0 | 94.2 | 0.0 |
| | **Overall** | 1.5 | 9.0 | 2.9 | 92.8 | 99.9 | 96.1 | 0.2 |
| | **Broiler** | 0.6 | 1.2 | 1.1 | 98.5 | 99.1 | 98.8 | 0.9 |
| **Conf50** | **Hen** | 0.9 | 2.8 | 1.7 | 95.7 | 99.1 | 97.4 | 0.9 |
| | **Overall** | 0.8 | 2.0 | 1.4 | 97.1 | 99.1 | 98.1 | 0.9 |
| | **Broiler** | 1.4 | 6.7 | 2.6 | 99.9 | 94.1 | 96.9 | 5.9 |
| **Conf75** | **Hen** | 1.8 | 12.5 | 3.5 | 99.6 | 89.6 | 94.3 | 10.4 |
| | **Overall** | 1.6 | 9.6 | 3.1 | 99.8 | 91.9 | 95.6 | 8.2 |

Where, MAE is the mean absolute error; MSE is the mean square error, RMSE is the root mean square error; FNR is the False negative rate; Conf12.5, Conf25, Conf50, and Conf75 are the confidence interval of 12.5%, 25%, 50%, and 75%, respectively.

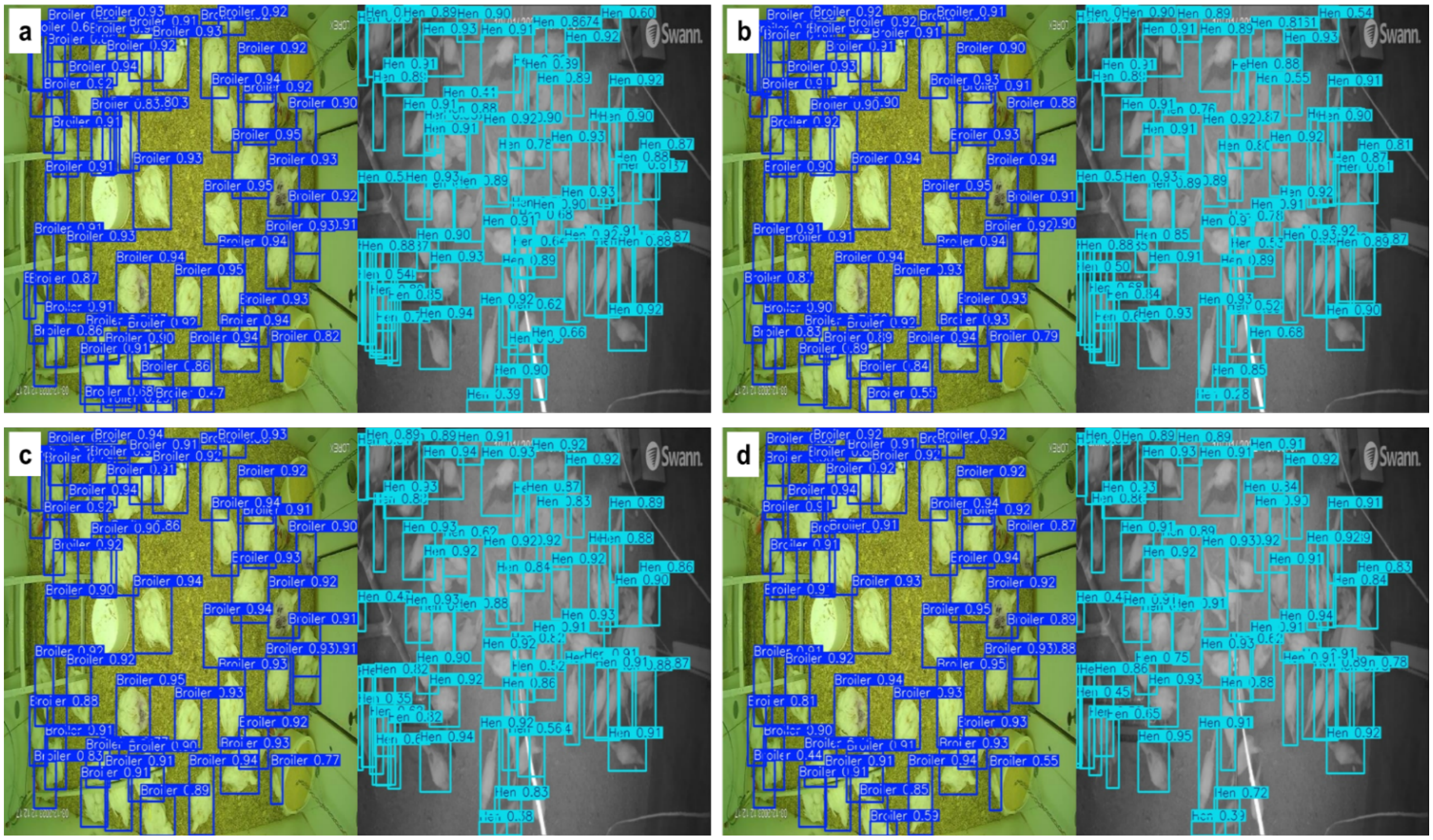

**Fig. 15.** Detected images of broilers and hens at different confidence levels: a) Conf12.5, b) Conf25, c) Conf50, and d) Conf75. The left side shows detected broilers, while the right side shows detected hens. Conf12.5, Conf25, Conf50, and Conf75 are the confidence intervals of 12.5%, 25%, 50%, and 75%, respectively.

High confidence intervals enable an uncertainty-aware pseudo-label assignment strategy, eliminating the need for manually designed thresholds and reducing noisy labels (Y. Wang et al., 2021). At the highest confidence level (75%), precision further improves for both broilers and hens, reaching near-perfect values. However, this comes at the cost of a significant drop in recall, especially for hens, which falls to around 90%. The overall performance at this level reflects this imbalance, with an increase in precision but a reduction in recall, leading to a slight decrease in the F1 score. Moreover, as the confidence levels rise, the false negative rate (FNR) also goes up. This means that although the model becomes more certain about its predictions, it tends to overlook some hens and broilers. Thus, the results highlight the challenge of balancing precision and recall, with higher confidence thresholds favoring precision at the expense of recall. It ultimately affects the F1 score and the model's ability to avoid false negatives.

*3.6. Performance comparison of manual and conf50 methods*

This study utilizes active learning to correct or label unlabeled or mislabeled images. Specifically, we compared two active learning approaches, the confidence threshold at 50% (conf50) and manual annotation with minimal human intervention, to determine which method yields better results. The confidence threshold was selected based on the comparison above, with conf50 yielding better detection accuracy. **Table 11** and **Fig. 16** compare broiler and hen detection performance metrics using the conf50 method and minimal manual correction. While the minimal manual correction method consistently demonstrated superior performance across all metrics, the

conf50 approach still showcased notable strengths, particularly its ability to achieve high precision and recall. For broilers, the conf50 method achieved a precision of 98.4%, recall of 97.7%, and an F1-score of 98.1%, indicating robust detection performance with automated thresholding. Similarly, for hens, the conf50 approach attained a commendable F1-score of 93.1% and a precision of 94.9%, highlighting its reliability in scenarios where minimal manual correction is infeasible.

**Table 11.** Validation performance metrics of manual and conf50 methods for broiler and hen detection across various models.

| Correction Method | Class | Precision | Recall | mAP_0.50 | mAP_0.50-0.95 | F1-score |
|---|---|---|---|---|---|---|
| **Conf50** | **Broiler** | 98.4 | 97.7 | 99.1 | 74.0 | 98.1 |
| | **Hen** | 94.9 | 91.3 | 96.5 | 72.1 | 93.1 |
| | **all** | 96.7 | 94.5 | 97.8 | 73.0 | 95.6 |
| **Manual** | **Broiler** | 99.6 | 99.1 | 99.4 | 80.8 | 99.4 |
| | **Hen** | 98.6 | 97.4 | 99.4 | 88.9 | 98.0 |
| | **all** | **99.1** | **98.3** | **99.4** | **84.9** | **98.7** |

Where mAP is the mean average precision, and Conf50 is the confidence interval of 50%. Manual and Conf50 are trained on 400 images of each broiler and hen.

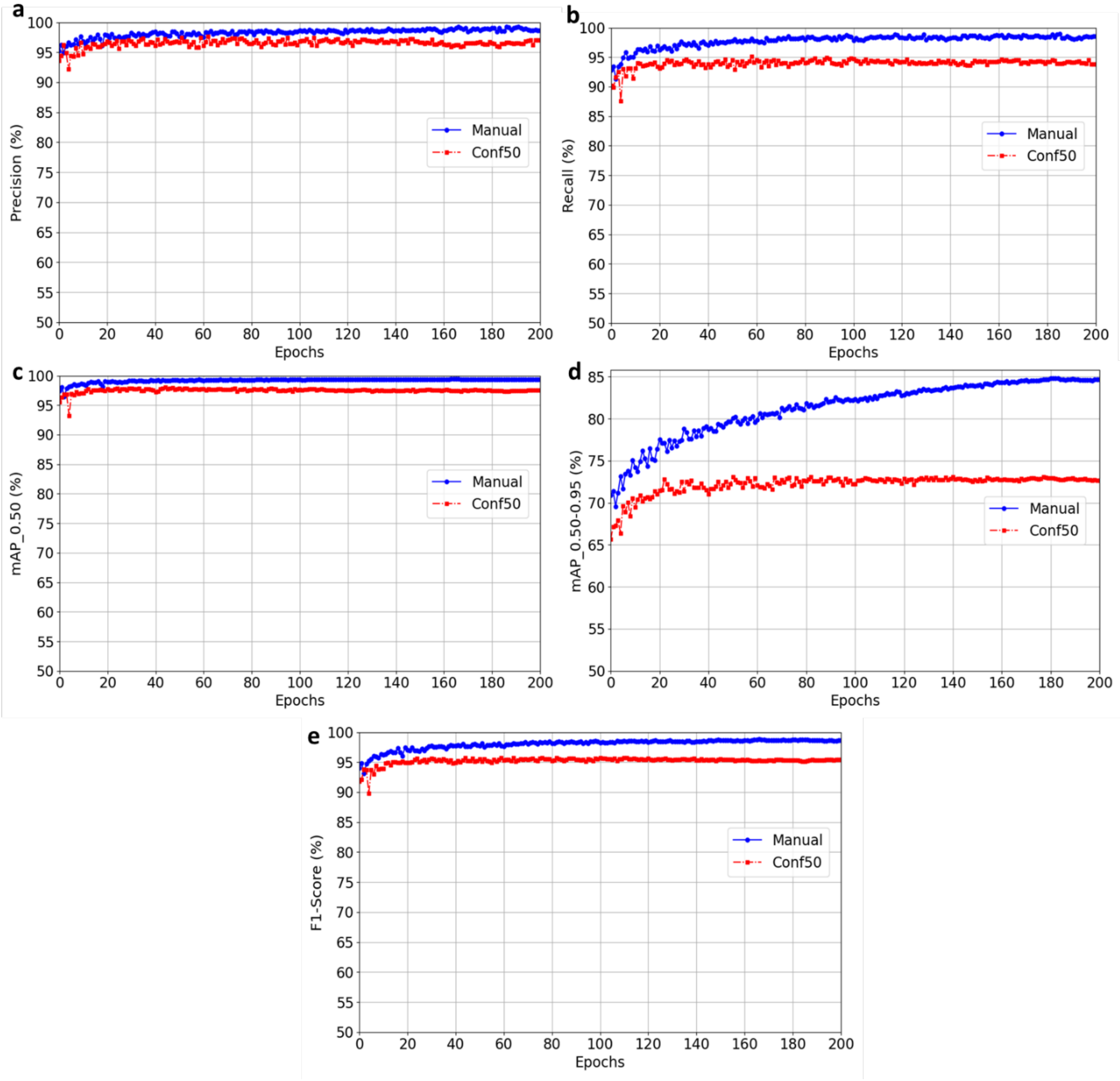

**Fig. 16.** Performance comparison of manual and conf50 methods for broiler and hen detection with a) precision, b) recall, c) mAP_0.50, d) mAP_0.50-0.95, and e) F1-score. conf50 – confidence level at 50% or more. mAP is mean average precision; Conf50 is confidence interval of 50%. Manual and Conf50 are trained on 400 images of each broiler and hen.

However, the manual correction method outperformed conf50 in refining detection accuracy, particularly in complex scenarios. It achieved a precision of 99.6%, a recall of 99.1%, an F1-score of 99.4% for broilers, and a mAP_0.50-0.95 of 80.8%, compared to conf50's 74.0%. For hens, the manual method also excelled with precision and an F1-score, alongside a notably higher mAP_0.50-0.95 versus conf50. Overall, the manual method yielded a combined F1-score of 98.7% and mAP_0.50-0.95 of 84.9%, significantly surpassing the conf50 values of 95.6% and 73.0%, respectively. Hence, minimal manual correction will be the best approach when resources are available. With this minimal effort, the model's robustness in accurately detecting can be improved.

The results presented in **Fig. 17** demonstrate the detection performance of the model for broilers, hens, and backgrounds under two evaluation approaches: conf50 and manual labeling. Under the conf50 threshold, the model successfully detected 1799 broilers and 1275 hens, but some false detections were noted, with 48 broilers and 107 hens incorrectly detected as background. Additionally, 31 broilers and 81 hens were misclassified under background predictions. When evaluated with manual labeling, the model exhibited enhanced detection accuracy, with 1820 broilers and 1342 hens detected correctly and only 15 broilers and 38 hens incorrectly labeled as background. The manual labeling approach also significantly reduced background detection errors, with only 10 broilers and 14 hens mislabeled as background. These results indicate that manual labeling provides a more accurate reflection of the model's detection capability while highlighting the need for further improvements in reducing false detections in background areas.

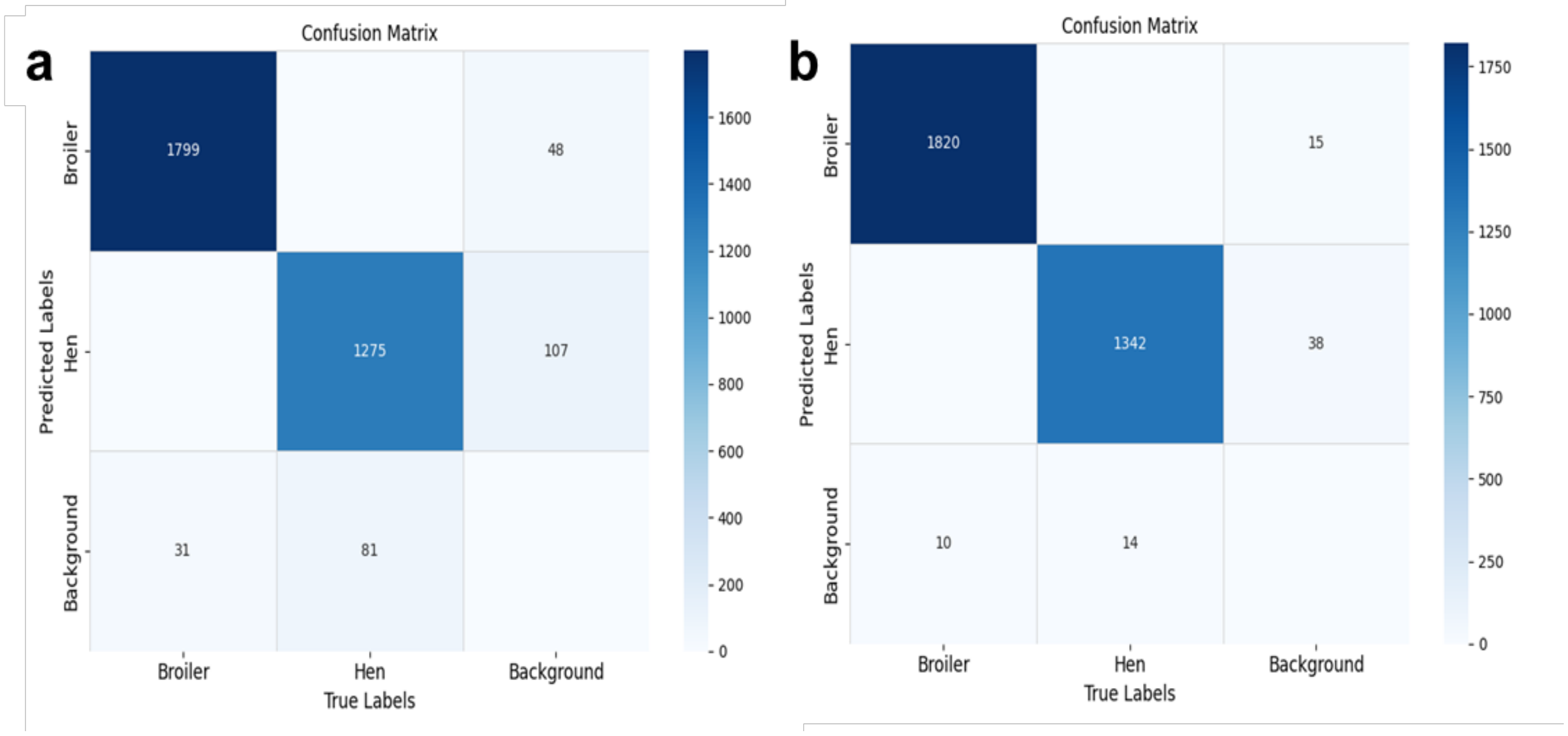

**Fig. 17.** Confusion matrix for hen and broiler detection: a) conf50 threshold and b) manual labeling. Conf50 is a confidence interval of 50%. Manual and Conf50 are trained on 400 images of each broiler and hen.

Similarly, based on the training and validation losses from **Fig.18**, comparing the Manual and Conf50 methods reveals distinct performance trends. Conf50 consistently outperforms Manual for training in terms of box_loss, cls_loss, and dfl_loss, showing more efficient optimization with lower values across all metrics. However, the Manual demonstrates a significant advantage when examining the validation losses, especially in box_loss, where it achieves a lower value than Conf50. This trend is also observed in the validation cls_loss and dfl_loss, where the Manual performs better. Despite Conf50's better training losses, its higher validation losses suggest that it may be overfitting the training data and struggling to generalize effectively to unseen data, like the behavior seen with Faster R-CNN in other comparisons. These results highlight the importance of balancing training and validation performance, as a model's superior training losses do not always translate to better generalization on validation and test datasets.

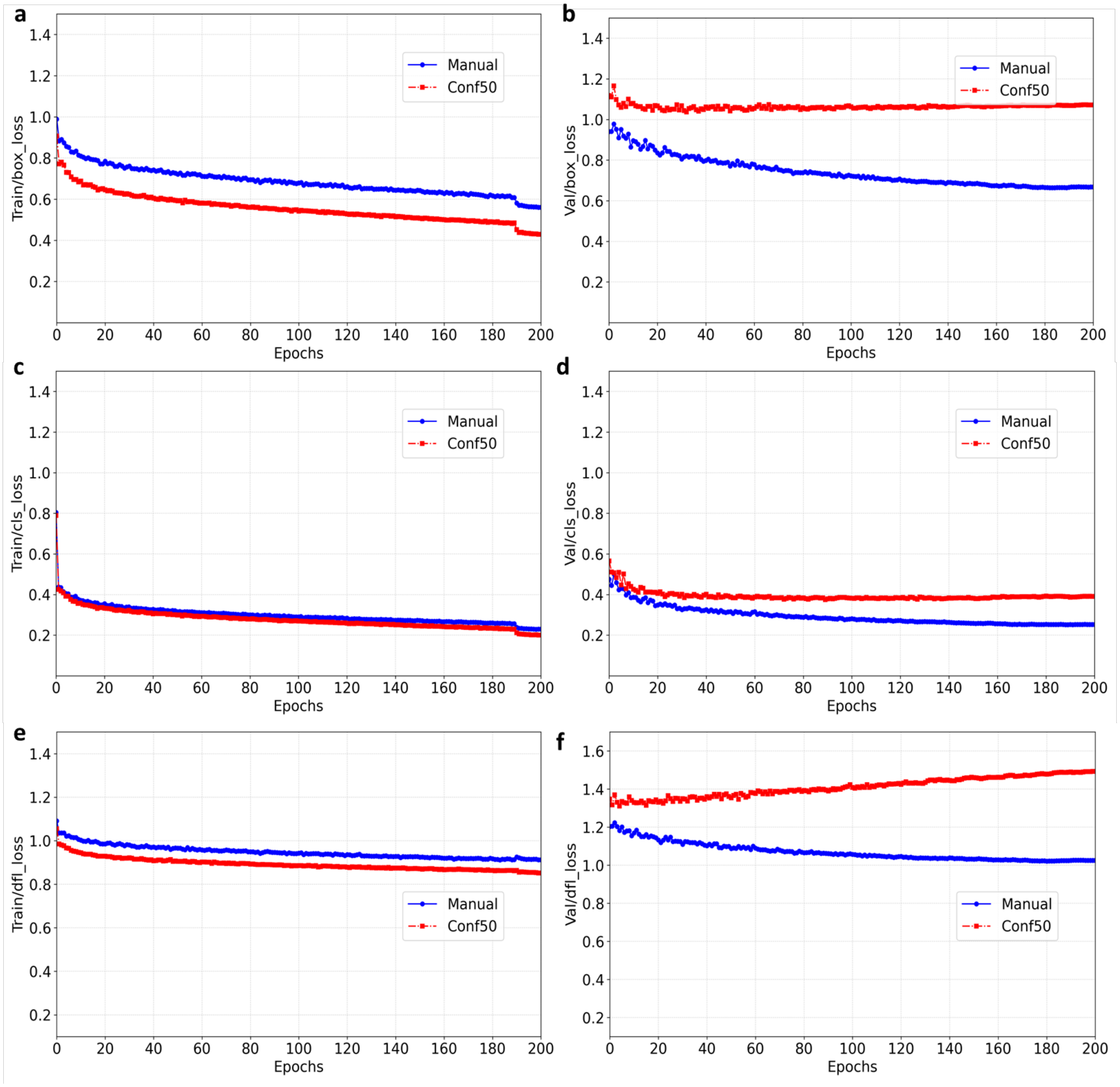

**Fig. 18.** Performance comparison of manual and conf50 methods for broiler and hen detection across a) Train box loss, b) Validation box loss, c) Train class loss, d) Validation class loss, e) Train dfl loss, and f) Validation dfl loss. Conf50 is a confidence interval of 50%; Val is validation; cls is class; dfl is distribution focal loss. Manual and Conf50 are trained on 400 images of each broiler and hen.

In addition, the test performance metrics for broiler and hen detection using the confidence threshold at 50% and manual correction methods, as shown in **Table 12**, highlight the clear superiority of the manual method in terms of accuracy. For broiler detection, the manual method significantly outperformed conf50, with an MAE of 0.26, MSE of 0.34, and RMSE of 0.58, all of which were notably lower than the corresponding values for conf50 (MAE = 0.55, MSE = 1.15, RMSE = 1.07). Additionally, the manual correction method achieved higher precision, recall, and F1-score than conf50 with a precision of 98.48%, recall of 99.13%, and F1-score of 98.8%. The FNR was also lower for the manual method (0.44%) compared to conf50 (0.87%), further emphasizing the improved accuracy and fewer missed detections with the manual correction. Unlike traditional methods that rely on manually defined thresholds, this strategy ensures a more accurate and efficient labeling process by automatically prioritizing high-confidence predictions.

**Table 12.** Prediction performance metrics of manual and conf50 methods for broiler and hen detection across various models.

| **Correction Method** | **Class** | **MAE** | **MSE** | **RMSE** | **Precision** | **Recall** | **F1 Score** | **FNR(%)** |
|---|---|---|---|---|---|---|---|---|
| **Conf50** | **Broiler** | 0.6 | 1.2 | 1.1 | 98.5 | 99.1 | 98.8 | 0.9 |
| | **Hen** | 0.9 | 2.8 | 1.7 | 95.7 | 99.1 | 97.4 | 0.9 |
| | **Overall** | 0.8 | 2.0 | 1.4 | 97.1 | 99.1 | 98.1 | 0.9 |
| **Manual** | **Broiler** | 0.3 | 0.3 | 0.6 | 99.3 | 99.6 | 99.4 | 0.4 |
| | **Hen** | 0.6 | 2.1 | 1.4 | 96.6 | 99.8 | 98.2 | 0.2 |
| | **Overall** | 0.5 | 1.2 | 1.0 | 98.0 | 99.7 | 98.8 | 0.3 |

Where MAE is the mean absolute error; MSE is the mean squared error; RMSE is the root mean squared error; FNR is the false negative rate; and Conf50 is the confidence threshold at 50%; Manual and Conf50 are trained on 400 images of each broiler and hen.

While the manual method consistently performed better, the Conf50 method still demonstrated competitive results for hen detection (**Fig. 19**), with precision of 95.73%, recall of 99.12%, and F1-score of 97.39%. The manual method again outperformed with precision, recall, and an F1-score, alongside a lower FNR of 0.22% compared to conf50's 0.88%. Despite these differences, the prediction results for the test datasets were quite similar. This suggests that a 50% confidence threshold could still improve accuracy, especially when manual labeling resources aren't available. Moreover, leveraging high confidence intervals allows for a dynamic pseudo-label assignment approach that adapts to uncertainty, effectively minimizing the inclusion of noisy labels (Wang et al., 2021). The conf50 method offers a reasonable alternative when speed and efficiency are prioritized, making it suitable for real-time applications with limited computational resources. However, if we have to choose the best method for model robustness, minimal manual annotation is the best approach.

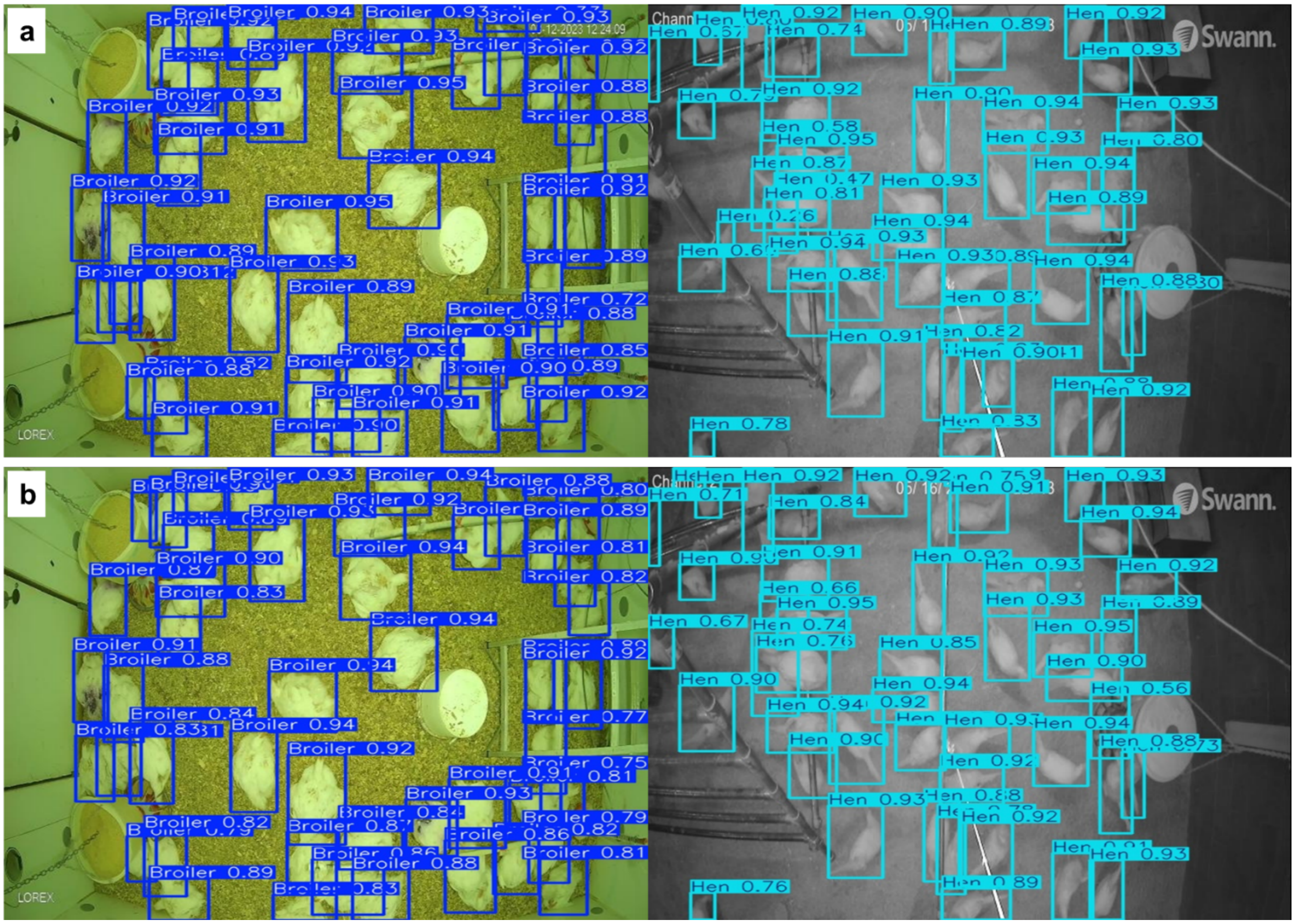

**Fig. 19.** Detected images of broilers and hens at a) Conf50 and b) Manual labeling methods. The left side shows detected broilers, while the right side shows detected hens.

*3.7. Human labeling, semi-supervised and unsupervised zero-shot auto labeling*

Labeling a dataset of 3,000 images containing 118,679 instances (approximately 50 instances per image) revealed significant differences in efficiency between human labeling and semi-supervised auto-labeling methods. Human annotation required 141.66 ± 12.25 seconds per image, totaling 118.05 ± 10.21 hours or nearly 5 days (4 days and 22 hours ± 10 hours 13 minutes) to complete the entire dataset. Assuming an uninterrupted 8-hour workday, excluding weekends, this process would require approximately 15 working days to complete. In contrast, semi-supervised auto-labeling alone processed all 3,000 images in only 2.4 ± 0.3 seconds, with an average of 0.0008 ± 0.0001 seconds per image, completing the task in mere seconds overall. For refinement, human intervention guided by a Query by Committee strategy added 28.65 ± 19.62 seconds per image, or 23.88 ± 16.35 hours in total. When accounting for an 8-hour workday, this hybrid method would require approximately 3 working days, achieving significant time savings over manual labeling while maintaining label quality.

Our results underscore the superiority of combining machine learning-driven labeling with targeted human input compared to previous work. As documented in earlier studies, fully manual methods are accurate but labor-intensive, requiring weeks, months, or years to annotate large datasets, making them impractical for rapid model development (Gudivada et al., 2017). On the other hand, unsupervised approaches, while faster, often compromise quality due to the lack of domain-specific corrections (Chen et al., 2021). Our semi-supervised approach bridges this gap by leveraging machine learning for rapid initial annotations and incorporating active learning methods (conf50 and minimal human expertise) for precise refinement. The hybrid method with minimal human intervention reduces annotation time by approximately 80% compared to manual labeling, completing the task in about 3 working days under the same conditions. This balance of speed, scalability, and accuracy establishes our method as an efficient and practical solution for annotating large-scale datasets in machine-learning workflows. A detailed performance evaluation of these labeling techniques, including their accuracy, precision, and other attributes, is clearly illustrated in **Table 13**, providing a comparative analysis of human, semi-supervised, and zero-shot auto-labeling techniques (Bist et al., 2023; Gudivada et al., 2017; Huang et al., 2014; Li et al., 2015; Nassar et al., 2021; Shorewala et al., 2021).

**Table 13.** Comparison of human labeling, semi-supervised auto-labeling, and zero-shot/unsupervised auto-labeling methods.

| Factor | Human Labeling | Semi-Supervised Auto-Labeling | Zero-shot or Unsupervised Auto-Labeling |
|---|---|---|---|
| **Accuracy & Precision** | High accuracy, but prone to human error and subjectivity. | Depending on the initial model's performance, it may be less accurate due to noisy labels. | Varies widely; can be inaccurate for complex tasks or rare classes. |
| **Consistency** | Varies between annotators; prone to subjective inconsistencies. | Generally, it is more consistent, especially for large datasets, but can suffer from model bias. | High consistency but can propagate errors if the initial clustering or segmentation is poor. |
| **Scalability** | Not scalable for large datasets; slow process. | Scalable can label massive datasets efficiently once the model is trained. | Highly scalable, as no manual intervention is needed for labeling. |
| **Cost-Effectiveness** | High labor costs, especially for large datasets. | Lower cost once the model is trained; initial setup may incur costs. | Very low cost, as no manual labeling is needed, but may require extensive computing resources. |
| **Label Quality & Reliability** | High-quality labels, particularly for complex tasks. | Label quality varies based on the model's confidence and initial training. | Quality can be poor, especially if the model is not properly trained or the data is noisy. |
| **Adaptability to New Data** | Human annotators can easily adapt to new classes or data. | It can adapt but may require retraining or fine-tuning to handle new data. | Limited adaptability to new classes may require clustering methods to handle new data. |
| **Handling Ambiguity & Uncertainty** | Humans can resolve ambiguity with contextual understanding. | May struggle with ambiguity; confidence thresholds can help reduce uncertainty. | Struggles with ambiguity, as unsupervised methods typically don't handle context well. |
| **Model Transferability & Improvement** | N/A (doesn't apply to human labeling). | Model improves over time with more data, leading to better pseudo-labels. | It can improve over time but requires fine-tuning or additional supervision for accuracy. |
| **Ease of Use** | Straightforward but time-consuming and labor-intensive. | Once set up, it is fast, efficient, and requires less manual input. | Fully automated; minimal manual input but requires good initial setup and computing resources. |
| **Reliability for Large Datasets** | Prone to errors with large-scale data. | Can reliably process large datasets with minimal human input. | Very reliable for large datasets, as no human input is required. |

| | | |
|---|---|---|
| | | However, errors may go unnoticed. |

*3.8.Limitations of this study*

Besides several challenges mentioned above, semi-supervised auto-labeling techniques in poultry datasets face further limitations that could impact their performance. One such limitation is the model's sensitivity to noise in the data, which can significantly affect the performance of the semi-supervised auto-labeling system. It can lead to inaccurate predictions and lower-quality labels. These variations in the training set can degrade the model's ability to generalize and correctly label new, unseen instances. To address this limitation, incorporating minimal manual annotation or exploring other ideas could be beneficial. Another critical limitation is the difficulty of incorporating domain-specific knowledge into the semi-supervised framework. Poultry datasets often contain subtle behavioral differences that may require expert interpretation to label correctly. The reliance on auto-labeling alone may not fully capture these nuances, potentially affecting the model's accuracy and its ability to distinguish between similar behaviors. Therefore, using domain adaptation and advanced models for semi-supervised learning could be one of the best choices to address these limitations.

Despite these challenges, semi-supervised auto-labeling remains an attractive option for large-scale poultry behavior detection, especially when paired with high-quality labeled data. However, as a previous study indicates, semi-supervised methods are prone to propagate errors from the initial labeling phase, which can result in a cascading effect on downstream predictions (Xu et al., 2023). Researchers demonstrated that while semi-supervised models can significantly reduce manual labeling efforts, they still require high-quality labeled data to perform optimally. In contrast, other studies on fully supervised learning methods have emphasized the importance of curated datasets for achieving robust model performance, but they are often limited by the time and labor needed to produce such datasets (Sapkota et al., 2024). Overall, these findings highlight the need for continued refinement of semi-supervised techniques to address issues of data quality, noise, and behavior variability in poultry datasets.

### *3.9.Future direction*

As noted, the effectiveness of deep learning applications is heavily reliant on large, diverse datasets that accurately reflect real-world conditions. Current research has highlighted the challenges associated with semi-supervised ALPD, particularly regarding noise. Therefore, future research should focus on improving data collection methodologies to enhance the dataset's diversity and accuracy, which could involve utilizing advanced data augmentation techniques and more robust data preprocessing to mitigate issues of noise and imbalance. Furthermore, semi-supervised models would benefit from being integrated with active learning strategies, where the model can selectively query labels on uncertain or ambiguous instances. A critical area of exploration should also be incorporating domain-specific knowledge into these models. Expert-driven label refinement could help address subtle behavioral differences in poultry, thus improving the precision of the auto-labeling process. Additionally, incorporating multi-modal data, such as combining visual, acoustic, and environmental data, may lead to better detection and classification of poultry behaviors, particularly in variable conditions.

Moreover, exploring their generalizability and scalability across different poultry breeds and farming conditions is crucial as semi-supervised learning techniques evolve. By focusing on more robust semi-supervised frameworks including domain adaptation that can adapt to changing environmental conditions, researchers could enhance the practical applicability of AI tools in poultry farming. This includes developing scalable systems that can handle large volumes of real-time data for decision-making in precision livestock farming. Research efforts should also explore integrating such models into automated monitoring systems that can provide continuous feedback to farmers and help to enable faster response times to health and behavioral issues in poultry. This advancement could lead to more efficient and sustainable poultry management practices. In addition, it also contributes to the overall goal of improving animal welfare and productivity in the agricultural sector.

## 4. Conclusions

This research highlights the potential of semi-supervised learning techniques to improve poultry behavior detection, especially in large datasets where manual labeling is both time-consuming and resource intensive. Combining human annotation with pseudo-labeling significantly boosts the speed and accuracy of data labeling. This approach can be particularly beneficial for large-scale operations looking to enhance their detection systems efficiently. Among the models tested, YOLOv8s-World and YOLOv9s performed exceptionally well under supervised learning conditions, with YOLOv8s-ALPD achieving the highest precision (96.1%) and recall (99%) among semi-supervised models. The CLIP zero-shot model showed promising results for breed detection, with a precision of 76.52% and recall of 64.06%. However, integrating zero-shot models with other models, such as YOLO-World with the YOLOv8s backbone, resulted in a text-prompt hybrid model for breed detection that delivered superior performance, with an RMSE of 0.99, precision of 99.2%, recall of 99.4%, and an F1 score of 98.7%. In addition, the minimal manual annotation active learning approach improved semi-supervised model robustness in detecting broilers and hens. This hybrid approach, coupled with human-in-the-loop annotation systems, greatly reduces the dependency on large, labeled datasets while improving detection accuracy. By merging semi-supervised learning, zero-shot models, and human annotation, this research presents an efficient, scalable solution for poultry monitoring, ultimately contributing to advancing precision livestock farming and more sustainable agricultural practices.

**Acknowledgments**

This project was supported by NSF Data Analytics that are Robust and Trusted (#1946391), USDA-NIFA AFRI (2023-68008-39853); Georgia Research Alliance (Venture Fund); The Egg Industry Center (EIC); University of Arkansas' Center for Food Animal Wellbeing

**CRediT authorship contribution statement**

**Ramesh Bahadur Bist:** Writing – review & editing, Writing – original draft, Investigation, Visualization, Validation, Supervision, Software, Methodology, Formal analysis, Data curation, Conceptualization. **Lilong Chai:** Writing – review & editing, Resources, Investigation, Methodology, Data curation, Conceptualization. **Shawna Weimer:** Writing – review & editing, Resources, Methodology, Data curation, Investigation. **Hannah Atungulu:** Data curation. **Chantel Pennicott:** Writing – review & editing, Investigation. **Xiao Yang:** Writing – review & editing, Investigation. **Chaitanya Pallerla:** Writing – review & editing, Investigation. **Yang Tian:** Writing – review & editing, Investigation. **Dongyi Wang:** Writing – review & editing, Writing – original draft, Investigation, Visualization, Validation, Software, Methodology, Formal analysis, Data curation, Conceptualization.

**Declaration of competing interest**

The authors declare that no competing financial interests or personal relationships could have influenced the work presented in this paper.

**Data availability**

Data will be made available on request

**References**

Adsule, A., Roy, S., Sharma, A. K., & Sengupta, S. (2024). SiamALNet: A Semi-supervised Siamese Neural Network with Active Learning Approach for Auto-Labeling. In A. K. Tripathi, D. Anand, & A. K. Nagar (Eds.), *Proceedings of World Conference on Artificial Intelligence: Advances and Applications* (pp. 257–269). Springer Nature. https://doi.org/10.1007/978-981-97-4496-1_20

Ahn, S., Kim, S., Ko, J., & Yun, S.-Y. (2023). Fine tuning pre trained models for robustness under noisy labels. *arXiv Preprint arXiv:2310.17668*.

Alzubaidi, L., Bai, J., Al-Sabaawi, A., Santamaría, J., Albahri, A. S., Al-dabbagh, B. S. N., Fadhel, M. A., Manoufali, M., Zhang, J., Al-Timemy, A. H., Duan, Y., Abdullah, A., Farhan, L., Lu, Y., Gupta, A., Albu, F., Abbosh, A., & Gu, Y. (2023). A survey on deep learning tools dealing with data scarcity: Definitions, challenges, solutions, tips, and applications. *Journal of Big Data*, *10*(1), 46. https://doi.org/10.1186/s40537-023-00727-2

Bist, R. B., Yang, X., Subedi, S., Bist, K., Paneru, B., Li, G., & Chai, L. (2024). An automatic method for scoring poultry footpad dermatitis with deep learning and thermal imaging. *Computers and Electronics in Agriculture*, *226*, 109481. https://doi.org/10.1016/j.compag.2024.109481

Bist, R. B., Yang, X., Subedi, S., & Chai, L. (2023). Mislaying behavior detection in cage-free hens with deep learning technologies. *Poultry Science*, 102729.

Boesch, G. (2024, October 5). *The Fundamental Guide to Faster R-CNN [2025]*. Viso.Ai. https://viso.ai/deep-learning/faster-r-cnn-2/

Cao, K. (2024). *Enhancing Machine Learning With Data-Efficient Methods*.

Chen, P., Li, L., Wu, J., Dong, W., & Shi, G. (2021). Unsupervised Curriculum Domain Adaptation for No-Reference Video Quality Assessment. *2021 IEEE/CVF International Conference on Computer Vision (ICCV)*, 5158–5167. https://doi.org/10.1109/ICCV48922.2021.00513

Chen, W., Lin, L., Yang, S., Xie, D., Pu, S., & Zhuang, Y. (2022). Self-supervised noisy label learning for source-free unsupervised domain adaptation. *2022 IEEE/RSJ International Conference on Intelligent Robots and Systems (IROS)*, 10185–10192.

Cheng, T., Song, L., Ge, Y., Liu, W., Wang, X., & Shan, Y. (2024). Yolo-world: Real-time open-vocabulary object detection. *Proceedings of the IEEE/CVF Conference on Computer Vision and Pattern Recognition*, 16901–16911.

Culman, M., Delalieux, S., Beusen, B., & Somers, B. (2023). Automatic labeling to overcome the limitations of deep learning in applications with insufficient training data: A case study on fruit detection in pear orchards. *Computers and Electronics in Agriculture*, *213*, 108196. https://doi.org/10.1016/j.compag.2023.108196

Djolonga, J., Yung, J., Tschannen, M., Romijnders, R., Beyer, L., Kolesnikov, A., Puigcerver, J., Minderer, M., D'Amour, A., & Moldovan, D. (2021). On robustness and transferability of convolutional neural networks. *Proceedings of the IEEE/CVF Conference on Computer Vision and Pattern Recognition*, 16458–16468.

Gudivada, V., Apon, A., & Ding, J. (2017). Data quality considerations for big data and machine learning: Going beyond data cleaning and transformations. *International Journal on Advances in Software*, *10*(1), 1–20.

Huang, G., Song, S., Gupta, J. N., & Wu, C. (2014). Semi-supervised and unsupervised extreme learning machines. *IEEE Transactions on Cybernetics*, *44*(12), 2405–2417.

Jabbar, H., & Khan, R. Z. (2015). Methods to avoid over-fitting and under-fitting in supervised machine learning (comparative study). *Computer Science, Communication and Instrumentation Devices*, *70*(10.3850), 978–981.

Kim, J., Min, Y., Kim, D., Lee, G., Seo, J., Ryoo, K., & Kim, S. (2022). *ConMatch: Semi-Supervised Learning with Confidence-Guided Consistency Regularization* (No. arXiv:2208.08631). arXiv. https://doi.org/10.48550/arXiv.2208.08631

Li, H., Rajbahadur, G. K., Lin, D., Bezemer, C.-P., Ming, Z., & Jiang. (2024). Keeping Deep Learning Models in Check: A History-Based Approach to Mitigate Overfitting. *IEEE Access*, *12*, 70676–70689. https://doi.org/10.1109/ACCESS.2024.3402543

Li, J., Chen, D., Qi, X., Li, Z., Huang, Y., Morris, D., & Tan, X. (2023). Label-efficient learning in agriculture: A comprehensive review. *Computers and Electronics in Agriculture*, *215*, 108412. https://doi.org/10.1016/j.compag.2023.108412

Li, X., Guo, Y., & Schuurmans, D. (2015). Semi-Supervised Zero-Shot Classification with Label Representation Learning. *2015 IEEE International Conference on Computer Vision (ICCV)*, 4211–4219. https://doi.org/10.1109/ICCV.2015.479

Liu, F., Tian, Y., Chen, Y., Liu, Y., Belagiannis, V., & Carneiro, G. (2022). ACPL: Anti-curriculum Pseudo-labelling for Semi-supervised Medical Image Classification. *2022*

*IEEE/CVF Conference on Computer Vision and Pattern Recognition (CVPR)*, 20665–20674. https://doi.org/10.1109/CVPR52688.2022.02004

Liu, S., Zeng, Z., Ren, T., Li, F., Zhang, H., Yang, J., Jiang, Q., Li, C., Yang, J., & Su, H. (2025). Grounding dino: Marrying dino with grounded pre-training for open-set object detection. *European Conference on Computer Vision*, 38–55.

Munir, M. A., Khan, M. H., Khan, S., & Khan, F. S. (2023). Bridging Precision and Confidence: A Train-Time Loss for Calibrating Object Detection. *2023 IEEE/CVF Conference on Computer Vision and Pattern Recognition (CVPR)*, 11474–11483. https://doi.org/10.1109/CVPR52729.2023.01104

Nassar, I., Herath, S., Abbasnejad, E., Buntine, W., & Haffari, G. (2021). All labels are not created equal: Enhancing semi-supervision via label grouping and co-training. *Proceedings of the IEEE/CVF Conference on Computer Vision and Pattern Recognition*, 7241–7250.

OpenAI. (2024). *Openai/CLIP* [Jupyter Notebook]. OpenAI. https://github.com/openai/CLIP (Original work published 2020)

Paneru, B., Bist, R., Yang, X., & Chai, L. (2024). Tracking dustbathing behavior of cage-free laying hens with machine vision technologies. *Poultry Science*, *103*(12), 104289. https://doi.org/10.1016/j.psj.2024.104289

Qazi, F., Naseem, M., Aslam, S., Attaria, Z., Jan, M. A., & Junaid, S. S. (2024). AnnoVate: Revolutionizing Data Annotation with Automated Labeling Technique. *VFAST Transactions on Software Engineering*, *12*(2), 24–30.

Ren, S., He, K., Girshick, R., & Sun, J. (2017). Faster R-CNN: Towards Real-Time Object Detection with Region Proposal Networks. *IEEE Transactions on Pattern Analysis and*

*Machine Intelligence*, *39*(6), 1137–1149. IEEE Transactions on Pattern Analysis and Machine Intelligence. https://doi.org/10.1109/TPAMI.2016.2577031

Sapkota, R., Paudel, A., & Karkee, M. (2024). *Zero-Shot Automatic Annotation and Instance Segmentation using LLM-Generated Datasets: Eliminating Field Imaging and Manual Annotation for Deep Learning Model Development* (No. arXiv:2411.11285). arXiv. https://doi.org/10.48550/arXiv.2411.11285

Shaikh, T. A., Rasool, T., & Lone, F. R. (2022). Towards leveraging the role of machine learning and artificial intelligence in precision agriculture and smart farming. *Computers and Electronics in Agriculture*, *198*, 107119.

Shen, A., Dai, M., Hu, J., Liang, Y., Wang, S., & Du, J. (2024). *Leveraging Semi-Supervised Learning to Enhance Data Mining for Image Classification under Limited Labeled Data* (No. arXiv:2411.18622). arXiv. https://doi.org/10.48550/arXiv.2411.18622

Shorewala, S., Ashfaque, A., Sidharth, R., & Verma, U. (2021). Weed Density and Distribution Estimation for Precision Agriculture Using Semi-Supervised Learning. *IEEE Access*, *9*, 27971–27986. IEEE Access. https://doi.org/10.1109/ACCESS.2021.3057912

Ultralytics. (2024). *Models Supported by Ultralytics*. https://docs.ultralytics.com/models

Usama, M., Qadir, J., Raza, A., Arif, H., Yau, K. A., Elkhatib, Y., Hussain, A., & Al-Fuqaha, A. (2019). Unsupervised Machine Learning for Networking: Techniques, Applications and Research Challenges. *IEEE Access*, *7*, 65579–65615. IEEE Access. https://doi.org/10.1109/ACCESS.2019.2916648

USDA. (2024). *Selected U.S. farm production expenses, 2023–24F*. http://www.ers.usda.gov/data-products/chart-gallery/gallery/chart-detail/?chartId=82241

Wang, X., Tang, J., & Whitty, M. (2022). Data-centric analysis of on-tree fruit detection: Experiments with deep learning. *Computers and Electronics in Agriculture*, *194*, 106748. https://doi.org/10.1016/j.compag.2022.106748

Wang, Y., Peng, J., & Zhang, Z. (2021). *Uncertainty-Aware Pseudo Label Refinery for Domain Adaptive Semantic Segmentation*. 9092–9101. https://openaccess.thecvf.com/content/ICCV2021/html/Wang_Uncertainty-Aware_Pseudo_Label_Refinery_for_Domain_Adaptive_Semantic_Segmentation_ICCV_2021_paper.html

Wolff, M., Brendel, W., & Wolff, S. (2023). The Independent Compositional Subspace Hypothesis for the Structure of CLIP's Last Layer. *ICLR 2023 Workshop on Mathematical and Empirical Understanding of Foundation Models*.

Xu, X., Zhou, F., Zhang, K., & Liu, S. (2023). CCGL: Contrastive Cascade Graph Learning. *IEEE Transactions on Knowledge and Data Engineering*, *35*(5), 4539–4554. IEEE Transactions on Knowledge and Data Engineering. https://doi.org/10.1109/TKDE.2022.3151829

Xu, Y., & Goodacre, R. (2018). On Splitting Training and Validation Set: A Comparative Study of Cross-Validation, Bootstrap and Systematic Sampling for Estimating the Generalization Performance of Supervised Learning. *Journal of Analysis and Testing*, *2*(3), 249–262. https://doi.org/10.1007/s41664-018-0068-2

Yang, J., Lu, J., Batra, D., & Parikh, D. (2017). *A faster pytorch implementation of faster r-cnn*.

Yang, X., Bist, R., Subedi, S., & Chai, L. (2023). A deep learning method for monitoring spatial distribution of cage-free hens. *Artificial Intelligence in Agriculture*.

Zhang, B., Wang, Y., Hou, W., Wu, H., Wang, J., Okumura, M., & Shinozaki, T. (2022). *FlexMatch: Boosting Semi-Supervised Learning with Curriculum Pseudo Labeling* (No. arXiv:2110.08263). arXiv. https://doi.org/10.48550/arXiv.2110.08263

Zhu, X., Vondrick, C., Ramanan, D., & Fowlkes, C. C. (2012). Do We Need More Training Data or Better Models for Object Detection?. *BMVC*, *3*(5).